\documentclass[lettersize,journal]{IEEEtran}
\newcommand{\eg}{\textit{e}.\textit{g}.}

\newcommand{\ie}{\textit{i}.\textit{e}.}

\usepackage{times} 
\usepackage[dvipsnames,table,svgnames]{xcolor}
\usepackage{graphicx}  
\usepackage{float} 
\usepackage{subfigure}
\usepackage{amsmath}
\usepackage{url}
\usepackage{multirow}
\usepackage{makecell, multirow, tabularx}
\usepackage{adjustbox}
\usepackage{tabularx}
\usepackage{xspace}
\usepackage{booktabs}
\usepackage{amssymb,mathrsfs,amsmath}
\usepackage{arydshln} 
\usepackage{amsfonts} 
\usepackage{algorithm}
\usepackage{algorithmic}
\usepackage{colortbl} 
\usepackage{xcolor} 
\usepackage[table]{xcolor}

\usepackage{tikz}
\usepackage{collcell}
\usepackage{hyperref}
\definecolor{hollywoodcerise}{RGB}{244, 0, 161}
\hypersetup{colorlinks,linkcolor={red},citecolor={hollywoodcerise},urlcolor={red}} 

\ifCLASSOPTIONcompsoc
  \usepackage[nocompress]{cite}
\else
  \usepackage{cite}
\fi

\def\BibTeX{{\rm B\kern-.05em{\sc i\kern-.025em b}\kern-.08em
    T\kern-.1667em\lower.7ex\hbox{E}\kern-.125emX}}
\usepackage{balance}
\begin{document}
\title{GoodSAM++: Bridging Domain and Capacity Gaps via Segment Anything Model for Panoramic Semantic Segmentation}

\author{Weiming Zhang~\IEEEmembership{Student Member,~IEEE,},Yexin Liu,~\IEEEmembership{Student Member,~IEEE,} Xu Zheng,~\IEEEmembership{Student Member,~IEEE,}
        Lin Wang*,~\IEEEmembership{Member,~IEEE,}\thanks{* Corresponding Author}
\thanks{Weiming Zhang is with the AI Thrust, HKUST(GZ), Guangdong, China. E-mail: Zweiming996@gmail.com. Yexin Liu is with the AI Thrust, HKUST(GZ), Guangdong, China. E-mail: yliu292@connect.hkust-gz.edu.cn. Xu Zheng is with the AI Thrust, HKUST(GZ), Guangdong, China. E-mail: zhengxu128@gmail.com.
Lin Wang is with AI/CMA Thrust, HKUST(GZ) and Dept. of CSE, HKUST, Hong Kong SAR, China, E-mail: linwang@ust.hk.
}
}

\markboth{Journal of \LaTeX\ Class Files,~Vol.~18, No.~9, September~2020}%
{How to Use the IEEEtran \LaTeX \ Templates}

\maketitle
\begin{abstract}
This paper presents \textbf{GoodSAM++}, a novel framework utilizing the powerful zero-shot instance segmentation capability of SAM (\ie, teacher) to learn a compact panoramic semantic segmentation model, \ie, student, without requiring any labeled data. GoodSAM++ addresses two critical challenges: 1) SAM's inability to provide semantic labels and inherent distortion problems of panoramic images;
2) the significant capacity disparity between SAM and the student. 
The `out-of-the-box' insight of GoodSAM++ is to introduce a teacher assistant (TA) to provide semantic information for SAM, integrated with SAM to obtain reliable pseudo semantic maps to bridge both domain and capacity gaps. 
To make this possible, we first propose a Distortion-Aware Rectification (DARv2) module to address the domain gap. It effectively mitigates the object deformation and distortion problem in panoramic images to obtain pseudo semantic maps.
We then introduce a Multi-level Knowledge Adaptation (MKA) module to efficiently transfer the semantic information from the TA and pseudo semantic maps to our compact student model, addressing the significant capacity gap.
We conduct extensive experiments on both outdoor and indoor benchmark datasets, showing that our GoodSAM++ achieves a remarkable performance improvement over the state-of-the-art (SOTA) domain adaptation methods. Moreover, diverse open-world scenarios demonstrate the generalization capacity of our GoodSAM++. Last but not least, our most lightweight student model achieves comparable performance to the SOTA models, \eg,~\cite{zheng2023look} with only 3.7 million parameters. Models and code can be found at \url{https://vlislab22.github.io/GoodSAM_Plus/}.

\end{abstract}

\begin{IEEEkeywords}
Panoramic Semantic Segmentation, Segment Anything, Knowledge Transfer
\end{IEEEkeywords}

\section{Introduction}

\IEEEPARstart{T}{he} rising demand for comprehensive omnidirectional scene understanding has driven the popularity of $360^\circ$ cameras. 
Compared with the pinhole cameras that only capture a limited field-of-view (FoV) while $360^\circ$ cameras provide an ultra-wide FoV of $360^\circ  \times 180^\circ$. 
The increasing fascination with $360^\circ$ cameras is reflected in their strong sensing capabilities for diverse learning systems and practical applications, such as autonomous vehicles ~\cite{wang2018self,zheng2023look,wang2020bifuse,jayasuriya2020active} and immersive experiences in augmented and virtual reality (AR/VR) devices~\cite{xu2018predicting,ai2022deep}.
This has inspired active research endeavors~\cite{zheng2023both,li2023sgat4pass,zhang2021transfer} for addressing scene understanding tasks, especially panoramic semantic segmentation.
This task offers comprehensive scene understanding by integrating perception with pixel-level predictions across $360^\circ$ data.
Generally, to seamlessly integrate spherical data with modern deep learning frameworks, the Equirectangular Projection (ERP) method is extensively utilized to represent the 360$^\circ$ data in 2D planar representation
\footnote{In this paper, omnidirectional and panoramic images are interchangeably used, and ERP images often indicate panoramic images.}, making it efficient in processing and analyzing panoramic images. 

\begin{figure}[t]
    \centering
    \includegraphics[width=\columnwidth]{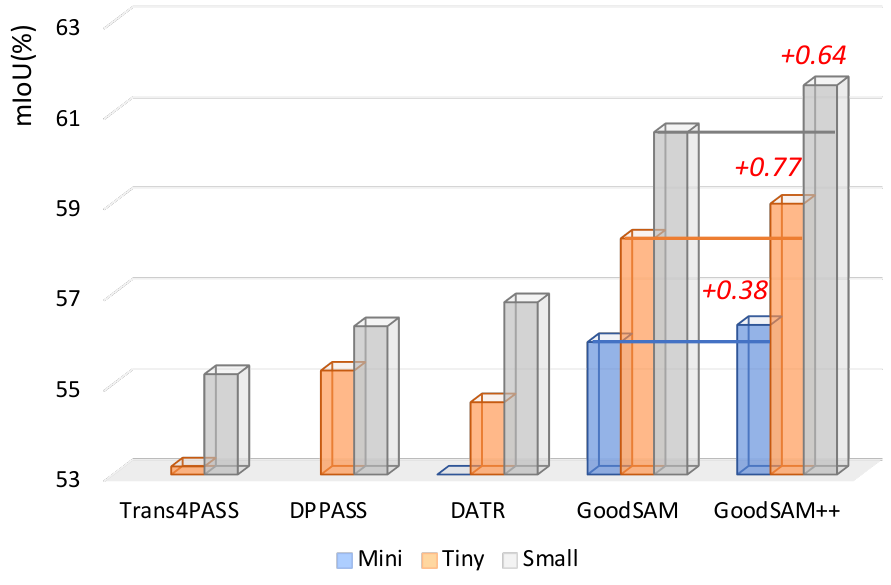}
    \caption{Performance comparison of GoodSAM++ with GoodSAM and previous SOTA methods ~\cite{zhang2022bending, zheng2023both, zheng2023look} across various model parameter ranges. Our GoodSAM++ outperforms our GoodSAM by the largest margin for the tiny size, with a performance gap of 0.77\% mIoU.
    }
    \label{fig:BEv2}
\end{figure}

However, ERP images often encounter issues such as large image distortions and object deformations due to the uneven distribution of pixels~\cite{ai2022deep,zheng2023look}. Consequently, numerous existing methods~\cite{xie2021segformer,yang2019pass,yang2020omnisupervised} tailored for pinhole images fall short in panoramic semantic segmentation, as they struggle to manage the significant distortions inherent in ERP images. Additionally, the limited availability of well-annotated datasets presents a significant challenge in training a robust foundation model for panoramic segmentation.

\begin{figure*}[t]
    \centering
    \includegraphics[width=\textwidth]{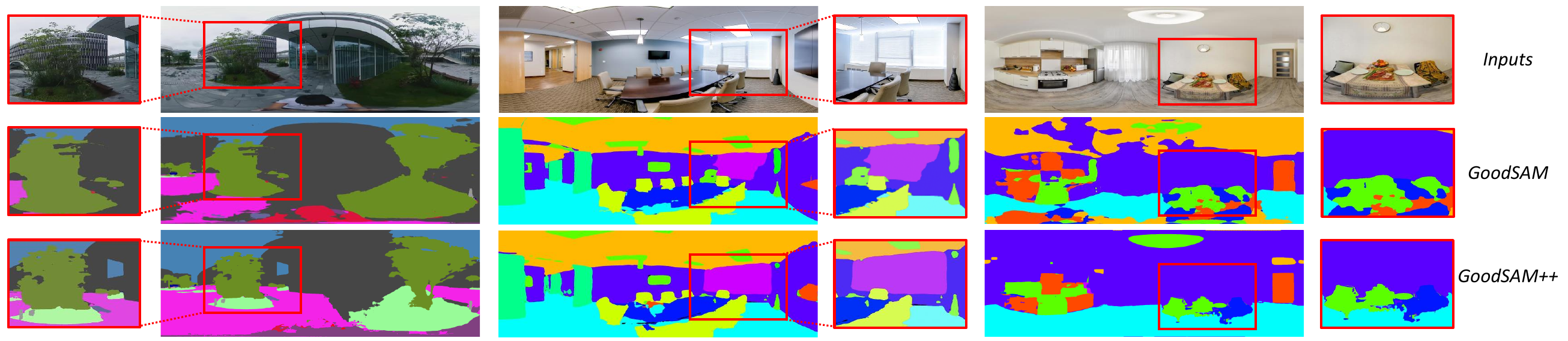}
    \caption{(a) Visual Comparisons between GoodSAM and GoodSAM++ on the diverse indoor or outdoor open-world scenes.
    }
    \label{teaser figure}
\end{figure*}

Therefore, researchers have explored transferring knowledge from the labeled pinhole image domain to the unlabeled panoramic image domain through unsupervised domain adaptation (UDA). ~\cite{liu2021pano,zhang2021transfer,zhang2022bending,zheng2023both,zheng2023look}. These methods can be generally categorized into three main groups.: pseudo labeling~\cite{liu2021pano, zhang2021deeppanocontext}, adversarial training~\cite{ma2021densepass, zheng2023both} and prototypical adaptation~\cite{zhang2022bending, zhang2022behind}. 
However, they often require many labeled 2D images for training or employ multi-branch designs~\cite{yang2020ds,zheng2023both}, leading to substantial computational costs.
Recently, the rapid development of foundational models has garnered widespread attention, demonstrating strong robustness across various tasks~\cite{kirillov2023segment,singh2022flava,nguyen2023lvm}. 
The Segment Anything Model (SAM)~\cite{kirillov2023segment}, one of the prominent vision foundation models, has been trained on large-scale datasets (over 1 billion masks on 11 million images). 
The large, diverse, and comprehensive training datasets enable SAM to perform exceptional zero-shot instance segmentation on unseen datasets. This strong capability makes SAM an ideal foundational model for many segmentation tasks across various domains~\cite{wu2023medical,ma2023segment,zhang2023segment}.


In this paper, we explore a novel question: \textit{How can we efficiently leverage SAM's powerful zero-shot instance segmentation capabilities to learn a compact panoramic semantic segmentation model, \ie, student, with unlabeled panoramic data?}
However, as SAM produces instance segmentation outputs without semantic information, directly transferring knowledge from SAM to our student model presents a task gap challenge. Additionally, the significant capacity gap between SAM and our lightweight student model makes task of learning a compact student model non-trivial.


To this end, we propose a novel framework, dubbed \textbf{GoodSAM++},
designed to learn a compact student model by leveraging SAM's powerful zero-shot capabilities. 
\textit{To obtain the semantic labels, a key insight of our GoodSAM++ framework is to introduce a Teacher Assistant (TA)}. It not only outputs semantic predictions that can be combined with SAM's outputs to generate pseudo semantic maps, but it also helps bridge the capacity gap between SAM and student model during the knowledge adaptation phase, thereby enhancing the efficiency of knowledge transfer. 

To obtain pseudo semantic maps and achieve efficient knowledge adaptation, GoodSAM++ enjoys two key technical contributions: Distortion-Aware Rectification (DARv2) module and Multi-level Knowledge Adaptation (MKA) module.
As previous version of the DAR module in GoodSAM~\cite{zhang2024goodsam} is overly reliant on TA's predictions for boundary pixels, and its performance is sub-optimal due to the multiple threshold settings.
Therefore, we propose a new module, called \textbf{DARv2} (Sec.~\ref{DAR}).
Specifically, DARv2 improves upon the DAR module in the three aforementioned aspects. 
We first adopt a more flexible window movement strategy, replacing the previous single horizontal movement. This increases prediction-level inconsistency in the overlapping areas of two windows, thereby enhancing the effectiveness of the consistency constraint.
Secondly, we propose a new version of the Boundary Enhancement block, named BEv2 block (Sec.~\ref{Sec.Boundary Enhancement Block}), to address the issue of TA's inaccurate predictions for boundary pixels. 
The BEv2 block is designed to incorporate the pseudo semantic maps’ boundary predictions from CTCF block to avoid the final refined boundary map missing boundary points due to TA boundary prediction omissions. 
Thirdly, to obtain higher quality pseudo semantic maps, we design a new CTCFv2 block (Sec.~\ref{Sec.CTCF}). This block adapts mask area evaluation standards and corresponding thresholds based on SAM's outputs, reducing the negative impact of manually-set thresholds. 
Upon obtaining reliable pseudo semantic maps, we introduce the MKA module to facilitate learning a compact student model (Sec.~\ref{MKA}). 
MKA supports knowledge transfer at multiple levels and scales, utilizing TA and pseudo-semantic maps to address both whole-image and window-based scales. Therefore, MKA minimizes the capacity gap between SAM and the student model, ultimately boosting the performance of our compact student model.

We conducted extensive experiments to validate our method. As shown in Fig.~\ref{teaser figure}(b),  our GoodSAM++ outperforms GoodSAM and SOTA UDA methods on two outdoor panoramic segmentation benchmarks across various model parameter ranges. Similarly, we also conducted experiments on a indoor dataset. The experiments demonstrated that in indoor scenarios, our GoodSAM++ also shows better performance compared to GoodSAM. Additionally, we used both online and our self-collected indoor/outdoor ERP images to test the generalization ability of GoodSAM++. Experimental results demonstrated that GoodSAM++ outperforms GoodSAM on open-world ERP images, showing a stronger zero-shot capability.

This work builds on our previous conference version~\cite{zhang2024goodsam} by introducing an enhanced model architecture and more comprehensive experimental evaluation. In summary, the additional contributions of this journal extension are as follows:
\begin{enumerate}
    \item To further mitigate the negative impact of distortion on TA and generate distortion-aware pseudo semantic maps, we propose the GoodSAM++ framework. Extensive experimental results (Tab.~\ref{perclass} and Fig.~\ref{visFigure}) demonstrate that our GoodSAM++ outperforms GoodSAM at various size levels of student models based on two outdoor benchmark detests.
    \item We propose a new DARv2 module (Sec.~\ref{DAR}) with novel designs in three aspects. 
    Specifically, 

    \begin{itemize}
        \item We introduce a more flexible sliding window strategy (Sec.~\ref{Sec.Consistency Constraint}), which enhances the effectiveness of the consistency constraint.
        \item Our improved BEv2 block (Sec.~\ref{Sec.Boundary Enhancement Block}) provides the TA with a more reliable pseudo-boundary map at the boundary level.
        \item In contrast, the CTCFv2 (Sec.~\ref{Sec.CTCF}) provides a reliable pseudo semantic map at the semantic level.
    \end{itemize}
    After optimization, DARv2 strengthens the connections between proposed blocks, improving its effectiveness in mitigating distortion for TA and enhancing the quality of pseudo semantic maps for our student model.

    \item We conduct new experiments (Fig.~\ref{visFigure_indoor}) on the \textit{indoor} panoramic semantic segmentation dataset, demonstrating superior performance of GoodSAM++  compared with GoodSAM and Trans4PASS+~\cite{zhang2024behind}.
    \item We validate the generalization ability of GoodSAM++ on diverse open-world and our self-collected indoor/outdoor data. The results (Fig.~\ref{visFigure_OPENINDOOR} \& \ref{visFigure_OPENOUTDOOR}) demonstrated that our GoodSAM++ exhibits superior generalization ability.
\end{enumerate}


\section{Related Work}
\noindent\textbf{Panoramic Image Semantic Segmentation.} 
The first line of works~\cite{yang2019pass, orhan2022semantic, yang2019can, xu2019semantic, yang2020ds, yang2020omnisupervised,zheng2024semantics,zheng2024360sfuda++,zhang2024behind,guttikonda2024single} on panoramic semantic segmentation are based on the supervised learning. However, due to the lack of comprehensive panoramic image datasets, the majority of current methods for panoramic image semantic segmentation rely on unsupervised domain adaptation (UDA)~\cite{zhu2023patch, zhu2023good}. Recent research endeavors have been focused on the UDA for panoramic semantic segmentation approaches, which can be divided into three types, including the pseudo labeling~\cite{liu2021pano, zhang2021deeppanocontext, zhang2017curriculum}, adversarial training~\cite{zhang2021transfer, ma2021densepass, zheng2023both} and prototypical adaptation~\cite{zhang2022bending, zhang2022behind, zheng2024semantics} methods. However, UDA methods require a substantial number of pinhole images for training, which significantly increases the associated training costs. \textit{Differently, we introduce SAM to the panoramic semantic segmentation task, aiming at transferring the instance segmentation knowledge of SAM to learn a compact student model without using any labeled data}.

\begin{figure*}[t]
    \centering
    \includegraphics[width=0.95\textwidth]{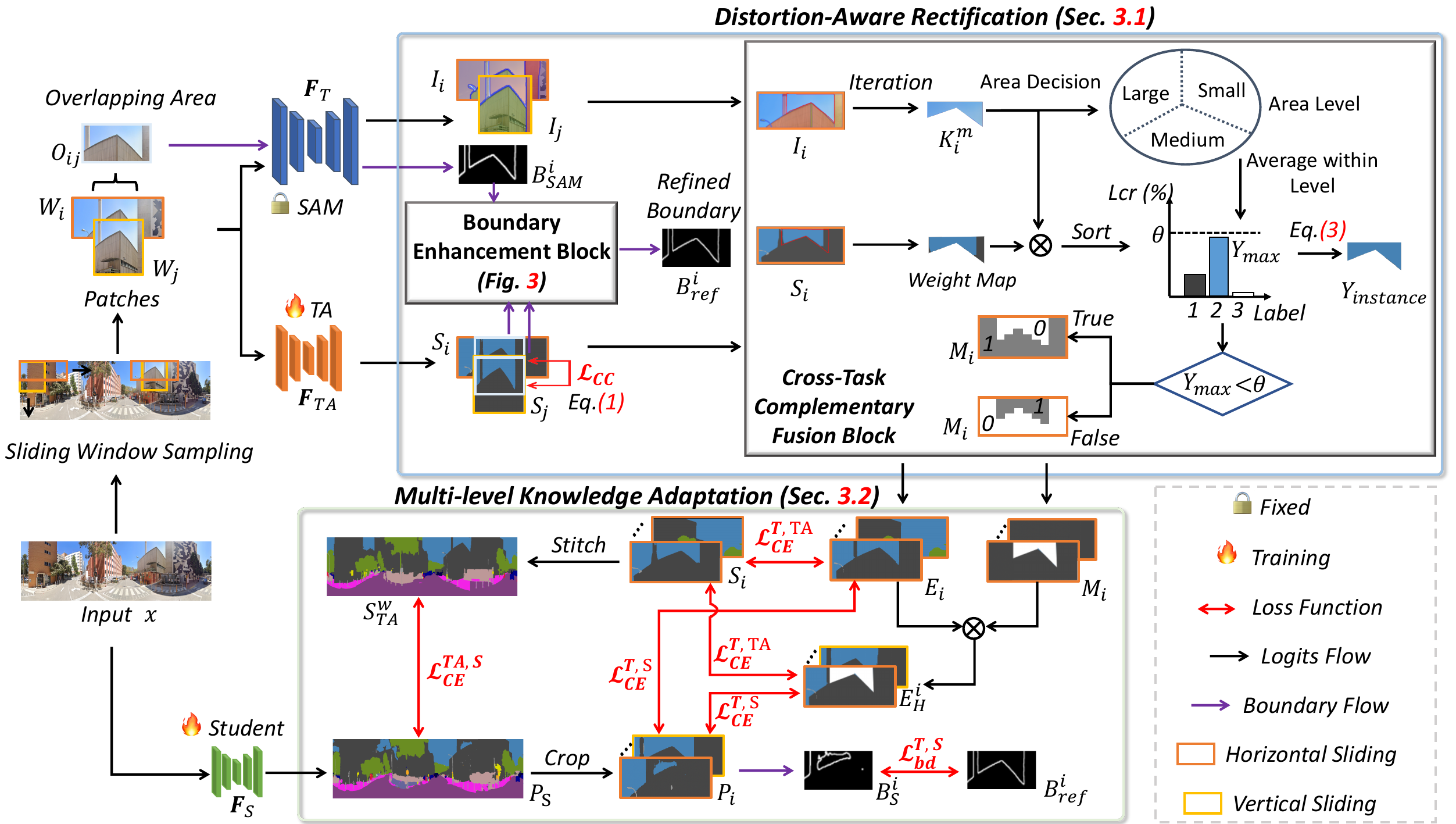}
    \caption{\textbf{Overview of GoodSAM++ framework}, consisting of three models: SAM, teacher assistant, and student. Our method has two main technical components: the Distortion-Aware Rectification (DAR) module and the Multi-level Knowledge Adaptation (MKA) module. 
    }
    \label{framework}
\end{figure*}

\noindent \textbf{Segment Anything Model (SAM)}
SAM serves as a foundational model for segmentation tasks~\cite{kirillov2023segment}. It leverages a dataset comprising 11 million diverse, high-resolution images and 1.1 billion meticulously annotated high-quality segmentation masks for training. This extensive dataset endows SAM with robust zero-shot instance segmentation capabilities. SAM has been applied to diverse tasks, such as medical image segmentation \cite{wu2023medical,ma2023segment,zhang2023segment,roy2023sam,huang2024segment,ma2024segment}, 3D segmentation \cite{cen2023segment,yang2023sam3d,zhou2024point,gong20233dsam}, image editing \cite{yu2023inpaint,zhang2023comprehensive}, and tracking \cite{cheng2023segment,yang2023track}. 
However, SAM has limitations in providing detailed semantic labels~\cite{kweon2024sam}. Additionally, the model's prohibitive number of parameters and the domain gap between its training data and panoramic images present significant challenges. These factors hinder SAM's ability to directly achieve accurate and effective panorama semantic segmentation. Consequently, additional strategies and adaptations are required to bridge these gaps and fully utilize SAM's capabilities in this task.
\textit{To address this, we introduce a TA model that integrates SAM's high-quality instance masks to generate reliable pseudo semantic maps, enabling us to learn a compact yet effective panoramic segmentation model.}

\noindent\textbf{Knowledge Transfer.}
Several methods enhance the performance of a student model by leveraging the expertise of multiple teachers and the similarities between their domains~\cite{ruder2017knowledge, he2019knowledge, ma2014knowledge,wang2021knowledge}. Knowledge adaptation involves transferring the most reliable and relevant knowledge from the teacher models to the student model, ensuring effective application in the target domain. However, current approaches face limitations, such as the risk of the student model over-fitting to the biases of the teacher models and difficulties in generalizing knowledge across different tasks or unseen data. 
Consequently, directly applying these existing methods to our panoramic segmentation task is impractical. \textit{To address this, we propose the CTCFv2 block, which adaptively combines the outputs from SAM and TA to generate more reliable pseudo semantic maps. Additionally, we introduce the MKA module, which utilizes TA's outputs and the pseudo semantic maps from the DARv2 module to perform multi-level and multi-scale knowledge transfer.} 

\section{Methodology}
\noindent\textbf{Overview.}
The overview of our GoodSAM++ framework is shown in Fig.~\ref{framework}. 
With the guidance from the SAM $\textbf{F}_T$ (\ie, teacher) and the assistance of the teacher assistant (TA) $\textbf{F}_{TA}$, our goal is to develop a compact panoramic semantic segmentation model (\ie, student) $\textbf{F}_S$ using the unlabeled panoramic images $x \in R^{H \times W \times 3}$.
Note that the TA is intended to provide essential semantic labels and to bridge the capacity gap between SAM and the student throughout the training process.
To mitigate the impact of ERP's large FoV, we adopt the overlapping sliding window strategy that moves windows of different sizes horizontally and vertically, to extract the local patches ${W_i}$ and ${W_j}$ from the input ERP images. 
As ${W_i}$ moves horizontally and ${W_j}$ moves vertically, we can obtain the overlapping area $O_{ij}$ between the two windows. To save computational cost, we set the movement stride in both horizontal and vertical directions equal to the window size.
Subsequently, the extracted patches (${W_i}$,${W_j}$) are input to both SAM $\textbf{F}_T$ and TA $\textbf{F}_{TA}$, yielding their respective predictions ($I_i$,$I_j$) and ($S_i$,$S_j$). Furthermore, the overlapping area $O_{ij}$ is exclusively input to $\textbf{F}_T$ to derive the corresponding boundary map ($B_{SAM}^i$). For the student model $\textbf{F}_S$, we input the ERP image $x$ to obtain the semantic prediction map $P_S$.
The challenges lie in:
1) effectively utilizing the predictions from SAM $\textbf{F}_T$ and TA $\textbf{F}_{TA}$ to obtain more reliable pseudo semantic maps $E_i$ and boundary predictions $B_{ref}^i$ as the supervision for student model $\textbf{F}_S$;
2) effectively performing knowledge adaptation from pseudo semantic maps $E_i$ and $\textbf{F}_{TA}$ to our compact student $\textbf{F}_S$.
To this end, we introduce the GoodSAM++ framework consisting of two key technical modules: the upgraded Distortion-Aware Rectification (DARv2) Module (Sec.~\ref{DAR}) and Multi-level Knowledge Adaptation (MKA) Module (Sec.~\ref{MKA}). We now describe these modules in detail.

\subsection{ Distortion-Aware Rectification (DARv2) Module}
\label{DAR}
The DARv2 module aims to obtain reliable supervision pseudo-labels by combining the predictions from SAM and TA. Due to the large FoV and distortion problems inherent in ERP images, which impact the performance of SAM and TA, we introduce an improved consistency constraint 
(CCv2), boundary enhancement (BEv2) block, and cross-task complementary fusion (CTCFv2) to obtain reliable pseudo semantic maps for panoramic images.
We now illustrate the details. 

\subsubsection{ Consistency Constraint}
\label{Sec.Consistency Constraint}
To enhance the performance of the TA on ERP images, we first introduce the prediction-level consistency constraint (CC) to help TA generate distortion-aware outputs. The original consistency constraint in the previous GoodSAM focuses on minimizing discrepancies in the overlapping areas present in the adjacent patches $W_i$ and $W_{i+1}$.
However, when two identical windows slide in the same direction, ensuring an overlapping region in the middle, the predictions in the overlapping area may differ slightly between the windows, but the gap between them is not significant.

\noindent \textbf{CCv2 block.}
Therefore, to enable the TA to more efficiently adapt to ERP images, we optimized the consistency constraint (CCv2) based on the new sliding window strategy. 
For the intersecting patches $W_i$ and $W_j$, due to distortion of ERP, $\textbf{F}_{TA}$'s predictions $S_i$ and $S_j$ for the overlapping area $O_{ij}$ between $W_i$ and $W_j$ exhibit more discrepancies. Therefore, we utilize the mean squared error (MSE) loss to ensure the consistency of predictions. As such, we can further enhance the $\textbf{F}_{TA}$'s sensitivity to local distortions compared with the previous version in GoodSAM. Formally, the consistency constraint loss $\mathcal{L}_{CCv2}$ is:

\begin{equation}
\setlength{\abovedisplayskip}{3pt}
\setlength{\belowdisplayskip}{3pt}
  \mathcal{L}_{CCv2}=MSE(S_i(O_{ij}),S_j(O_{ij})),
  \label{eq:W}   
\end{equation}

where the $S_i(O_{ij})$ denotes the TA's prediction within the overlapping area for the $(i,j)$ th window.



\subsubsection{Boundary Enhancement Block}
\label{Sec.Boundary Enhancement Block}
Due to SAM's strong zero-shot capability to provide relatively accurate boundary maps, we propose a boundary enhancement block \textit{to refine boundary pixels in TA's predictions}, inspired by \cite{rong2023boundary}. 
By improving TA's accuracy in predicting boundary pixels, we aim to alleviate the impact of distortion and object deformation on TA.
This block comprises two components: boundary refinement strategy and boundary-enhanced loss.
For two intersecting windows $W_i$ and $W_j$, we obtain two separate boundary maps $B_{TA}^i$ and $B_{TA}^j$ of the overlapping area $O_{ij}$. 
The boundary refinement strategy is proposed to identify reliable boundary pixels within the overlapping area by combining $B_{TA}^i$, $B_{TA}^j$, and $B_{SAM}^i$ of SAM to obtain the refined boundary map $B_{ref}^i$.   

The detailed Algorithm and description for the previous BE block design can be found in our conference version~\cite{zhang2024goodsam}.
Specifically, for the input boundary map $B_{TA}^i$, we first iterate through its boundary pixels and find corresponding pixels in $B_{TA}^j$ and $B_{SAM}^i$ at the same positions. 
If the pixels at the corresponding positions are all located on the boundary, they are deemed reliable boundary pixels.
For cases where this condition is not met, we search for the corresponding pixels in $B_{SAM}^i$ at the same positions and identify the nearest boundary pixel along the vertical axis.

Subsequently, we locate pixels at the same positions in $B_{TA}^i$ and $B_{TA}^j$, and for each pixel, perform softmax on its logits ($1\times 1 \times C$, where $C$ is the number of categories). 
We calculate the difference in the top two softmax values for each corresponding pixel in $B_{TA}^i$ and $B_{TA}^j$, denoting them as $D_i$ and $D_j$ respectively.
When either $D_i$ or $D_j$ has a value less than $\alpha$, we determine that the boundary pixel from $B_{SAM}^i$ exhibits the characteristics of the boundary pixel in the other two boundary maps as well. 
Thus, we define this boundary pixel in the $B_{SAM}^i$ as reliable. 
The parameter $\alpha$ determines the influence strength of SAM boundary pixels. Finally, if none of the above conditions are met, we decide to retain the boundary pixels of $B_{TA}^i$ as reliable pixels.
This way, we attain a refined boundary map $B_{ref}^i$ with all reliable boundary pixels for $O_{ij}$, which is utilized for updating TA $\textbf{F}_{TA}$ and student $\textbf{F}_{S}$.
\begin{figure}[t]
    \centering
    \includegraphics[width=\columnwidth]{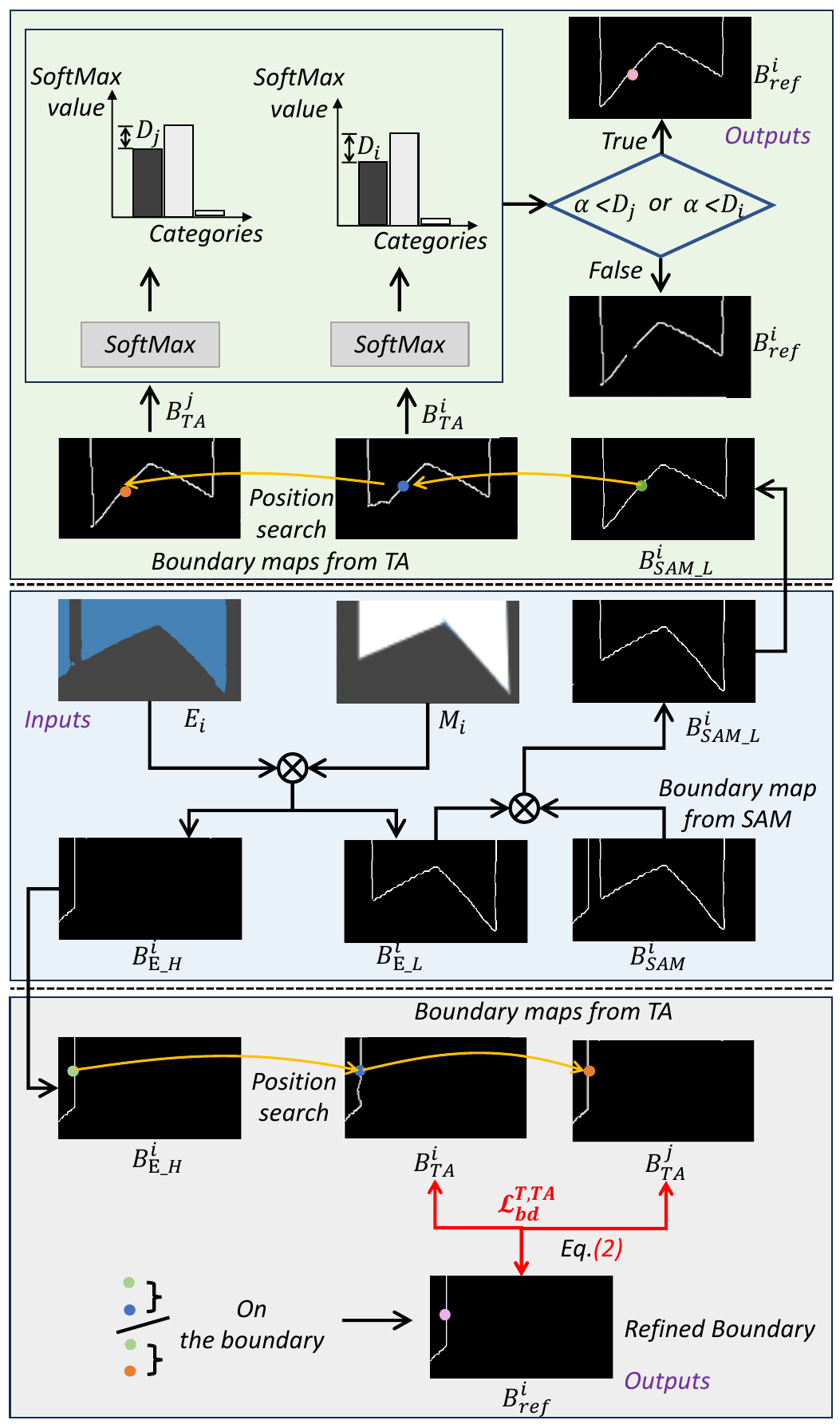}
    \caption{\textbf{Overview of the proposed BEv2 block}. In blue shaded part, by combining pseudo semantic maps, the confidence map, and the boundary map generated by SAM, we obtain the high and low confidence boundary maps. In  the bottom part, it represents the processing of pixels in the high confidence boundary map $B_{E\_H}^i$. Additionally, the green shaded part demonstrates the consideration of whether the pixels in the low confidence boundary map $B_{SAM\_L}^i$ are reliable.
    }
    \label{fig:BEv2}
\end{figure}

\noindent \textbf{BEv2 Block.}
However, we find that although the above BE algorithm overly relies on the performance of $B_{TA}^i$. Once the boundary pixels predicted in $B_{TA}^i$ are missing or contain numerous errors due to noise, the performance of $B_{ref}^i$ will degrade because the BE block traverses $B_{TA}^i$.
Therefore, to further enhance the accuracy of boundary predictions in $B_{ref}^i$, we propose the updated boundary enhancement block (BEv2) as illustrated in Fig.~\ref{fig:BEv2}. BEv2 incorporates additional pseudo semantic maps $E_i$ and a confidence map $M_i$ from the CTCF block (refer to Sec.~\ref{Sec.CTCF} ) to classify the boundary pixels into high or low confidence levels, and considers them differently to obtain a more accurate boundary map.

Specifically, as shown in Fig.~\ref{fig:BEv2} (a), we first fuse the pseudo semantic maps $E_i$ and the confidence map $M_i$ to obtain the high confidence boundary map $B_{E\_H}^i$ in the high confidence regions. For regions composed of the low confidence mask, we obtain the low confidence boundary map $B_{SAM\_L}^i$ by combining it with the SAM's boundary map $B_{SAM}^i$. 
Then, we traverse the boundary pixels (denote as green point) in $B_{E\_H}^i$ (See Fig.~\ref{fig:BEv2} (b)). If the green point and the corresponding blue point in $B_{TA}^i$ or the orange point in $B_{TA}^j$ are on the boundary, it indicates that the pixel represented by the green point is a reliable boundary pixel. If the above condition is not met, the green point is added to the $B_{SAM\_L}^i$.
Next, we analyze the boundary pixels in $B_{SAM\_L}^i$. The specific process remains consistent with the previous boundary enhancement block. It can be observed that the general process in Fig.~\ref{fig:BEv2}  is similar, with the only change being that when $D_i$ and $D_j$ are both greater than $\alpha$, we choose to discard this boundary pixel.
Through these steps, BEv2 constructs a refined boundary map $B_{ref}^i$ for the overlapping area $O_{ij}$, ensuring the inclusion of reliable boundary pixels.

Next, we introduce a boundary-enhanced loss (See Fig.~\ref{fig:BEv2}) to encourage TA's boundary pixel predictions to align closely with the refined boundary map: 

\begin{equation}
\setlength{\abovedisplayskip}{3pt}
\setlength{\belowdisplayskip}{3pt}
\begin{aligned}
\mathcal{L}_{bd}^{T,TA}=\sum_{k = 1}^{H\times W}\frac{(|B_{ref}^i-B_{TA}^i|+|B_{ref}^i-B_{TA}^j|)}{C_o},
\label{eq:boundary loss of T and TA}
\end{aligned} 
\end{equation}

where $C_o$ denotes the total boundary pixel counts of $B_{ref}$ and $k$ denotes the $k$-th pixel in the boundary map. This explicitly mitigates boundary pixel prediction errors caused by the distortion of ERP. 

\subsubsection{Cross-Task Complementary Fusion (CTCF).}
\label{Sec.CTCF}
To obtain more reliable pseudo semantic maps $E_i$ for window-based regions, we propose the CTCFv2 block, as shown in Fig.~\ref{framework}. It adaptively fuses SAM $\textbf{F}_{T}$'s instance mask outputs $I_i$ with TA $\textbf{F}_{TA}$'s semantic segmentation outputs $S_i$.
The objective of the fusion is to assign the highest-confidence semantic label to each instance mask based on the logits $S_i$ from TA (See Fig.~\ref{framework}).
Different from directly finding the most frequent or area-dominant semantic label~\cite{chen2023semantic,chen2023segment}, we define distinct area proportion thresholds for masks of different sizes and adaptively select the most reliable label.
Specifically, for each instance mask $K_i^m$ from SAM, we assess the instance mask area. For larger and smaller instance masks, we set the threshold $\theta$ to a smaller value to facilitate obtaining the most frequently occurring label. For medium-sized masks, we increase $\theta$ to a larger value to ensure the acquisition of a more accurate semantic label. 
Then, we identify the top three semantic labels in descending order of quantity within the corresponding area in TA's predictions. 
If the label coverage rate \textit{(lcr}) of the most prevalent semantic label $Y_{max}$ exceeds the $\theta $, we directly assign the most prevalent label as the semantic label for the current instance mask $K_i^m$. If the coverage rate of the most prevalent label falls below $\theta $, we delve into the three semantic labels and calculate their Shannon entropy (SE) using the logits $S_i$ from TA. The label with the minimum entropy is chosen as the highest-confidence semantic label $Y_{instance}$ for the $K_i^m$. The formulation is as follows:

\begin{equation}
\setlength{\abovedisplayskip}{3pt}
\setlength{\belowdisplayskip}{3pt}
Y_{instance}=\left\{
\begin{array}{rcl}
 Y_{max},& & {lcr(Y_{max})\geq \theta},\\
\\
Y_{argmin\{SE(Y_a)\}}, & & {0 <{lcr(Y_a)}< \theta},
\end{array} \right.
\end{equation}

where $a$ belongs to the top three semantic labels occupying the instance mask. 

\noindent \textbf{CTCFv2 Block.}
However, for different images, setting the same area range threshold and label coverage threshold $\theta $ can lead to significant human interference and fusion bias. Therefore, we optimize the CTCF block to obtain the CTCFv2 block. The detailed pseudo-code can be found in the Alg.~\ref{alg：fusion mechanism}
The CTCFv2 block first divides all instance masks into three area levels based on their sizes based on K-means~\cite{krishna1999genetic} algorithm. Then, for each area level, it averages the $Y_{max}$ of the masks within that level to obtain the corresponding threshold $\theta $ corresponding to that area level. Subsequently, during the fusion phase, CTCFv2 block follows the same fusion logic as the original CTCF block.
Through the CTCFv2 block, DAR produces high-quality pseudo semantic maps by adaptively merging the predictions based on windows from both SAM and TA.

\begin{algorithm}[t!]
 \caption{CTCFv2 Fusion Mechanism} 
 \label{alg：fusion mechanism} 
 \begin{algorithmic}[1]
     \STATE \textbf{Input}: The overlapping area instance masks $I_i$ from SAM. The overlapping area semantic map $S_i$ from TA.
     \STATE \textbf{Output}: pseudo semantic maps $E_i$ for the overlapping area.
     \STATE Sort all instance masks in $I_i$ by area in descending order.
     \STATE Apply K-means algorithm to divide instance masks into three size levels: Large, Medium, Small.
     \STATE Calculate the most prevalent semantic label proportion average value for each size level as $\theta_{Large}$, $\theta_{Medium}$, $\theta_{Small}$.
     \FOR{Each instance mask $K_i^m$ in $I_i$}
     \STATE Find the same region in $S_i$ as $K_{Sem}$ corresponds to the instance mask $K_i^m$.
     \STATE Calculate the count of each label in $K_{Sem}$ and sort them in descending order.
     \STATE Identify the top three labels $Y_a$ (a belongs to the number of categories) in the sorted order.
     \IF{$K_i^m$ belongs to Large}
             \STATE Use $\theta_{Large}$ for comparison.
         \ELSIF{$K_i^m$ belongs to Medium}
             \STATE Use $\theta_{Medium}$ for comparison.
         \ELSIF{$K_i^m$ belongs to Small}
             \STATE Use $\theta_{Small}$ for comparison.
         \ENDIF
     \IF{the lcr of most prevalent semantic label $Y_{max} > \theta$}
     \STATE $Y_{instance} = Y_{max}$.
     \ELSE
     \STATE Calculate the Shannon entropy value of $Y_a$ based on $S_i$.
     \STATE Find the $Y_{argmin\{SE(Y_a)\}}$ which the SE value is the smallest.
     \STATE $Y_{instance} = Y_{argmin\{SE(Y_a)\}}$.
     \ENDIF
     \ENDFOR
 \end{algorithmic} 
\end{algorithm}

Due to the fusion process potentially resulting in varying prediction confidences among masks, we obtain a weight map ($M_i$) with spatial dimensions identical to $W_i$ in the fusion process. Specifically, we assign higher weight (1) to masks with higher overlap between instance masks and semantic logits and lower weight (0) to masks that require SE for label assignment.
The pseudo semantic maps and weight maps obtained by the CTCFv2 block can assist the TA and the student in achieving better supervision for patch predictions.


\begin{figure*}[t!]
    \centering
    \includegraphics[width=\textwidth]{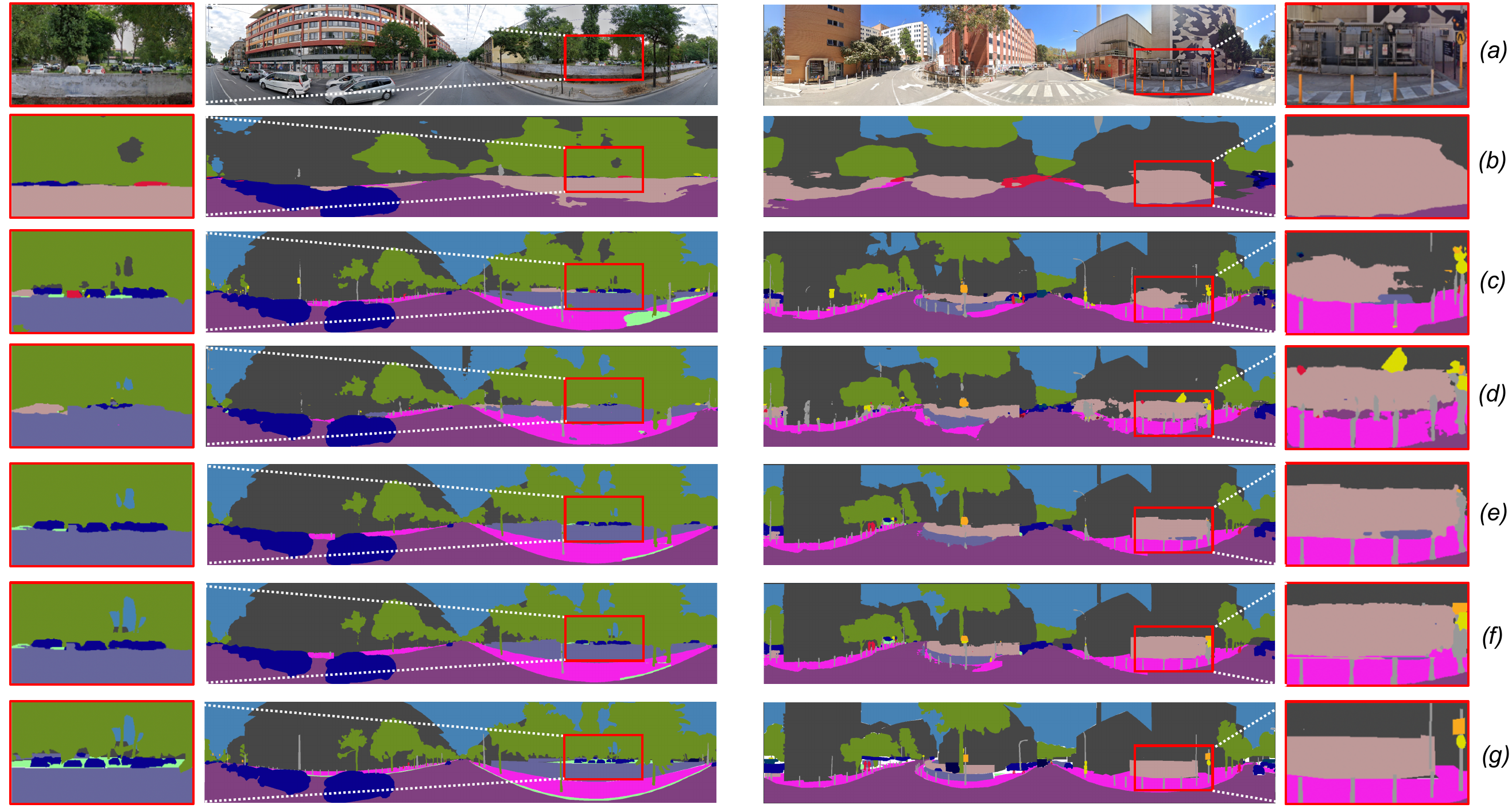}
    \caption{Example visualization results from the DensePASS test set: (a) Input panorama image, (b) Segformer-B5~\cite{xie2021segformer} without sliding window sampling, (c) DPPASS-S~\cite{zheng2023both}, (d) DATR-S~\cite{zheng2023look}, (e) GoodSAM-S, (f) GoodSAM++-S, (g) Ground truth.}
    \label{visFigure}
\end{figure*}

\subsection{Multi-level Knowledge Adaptation Module}
\label{MKA}
After addressing the distortion problem with our DARv2 module, we then propose the MKA module to learn a compact student module with TA $\textbf{F}_{TA}$'s output logits $S_i$  and the pseudo semantic maps $E_i$ from the DARv2 module for multi-level, multi-scale (whole-image and patch) knowledge adaptation.
To effectively transfer knowledge from the whole-image scale prediction of the TA $\textbf{F}_{TA}$ to the student model $\textbf{F}_{S}$, we concatenate predictions of non-overlapping patches to generate the entire ERP semantic prediction map $S_{TA}^{w}$.  
Therefore, when the entire image is directly fed to the student model $\textbf{F}_{S}$, resulting in the prediction map $P_{S}^{w}$. We use the Cross-Entropy (CE) loss (Eq.(\ref{eq:ce(TA,S)})) to guide the student in aligning its predictions with $S_{TA}^{w}$ at whole-image scale:. 

\begin{equation}
\setlength{\abovedisplayskip}{5pt}
\setlength{\belowdisplayskip}{5pt}
  \mathcal{L}_{ce}^{TA,S}= \mathcal{L}_{CE}(S_{TA}^{w},P_{S}^{w}).
  \label{eq:ce(TA,S)}   
\end{equation}


For the obtained pseudo semantic maps $E_i$ based on the patch $W_i$, we use another CE loss to guide the student's prediction logits at the corresponding window position $P_{S}^{i}$ to mimic $E_i$. 
As the CTCF block returns the weight maps $M_i$ corresponding to the higher confidence masks based on the fusion mechanism, we first combine weight maps with pseudo semantic maps to obtain higher confidence masks $E_H^i$. Then we perform knowledge adaptation simultaneously using pseudo semantic maps $E_i$ and higher confidence masks $E_H^i$.
Therefore, the loss for patches knowledge from $E_i$ and $E_H^i$ transfer to the student $\textbf{F}_{S}$ and TA $\textbf{F}_{TA}$ can be formulated as:

\begin{equation}
  \mathcal{L}_{CE}^{T,S}=\sum_{k=1}^{H\times W}(\mathcal{L}_{CE}(P_{i}^k,E_i^k)+\lambda M_i\mathcal{L}_{CE}(P_{i}^k,E_i^k)),
  \label{eq:ce(TA,S)}   
\end{equation}

\begin{equation}
  \mathcal{L}_{CE}^{T,TA}=\sum_{k=1}^{H\times W}(\mathcal{L}_{CE}(S_{i}^k,E_i^k)+\lambda M_i\mathcal{L}_{CE}(S_{i}^k,E_i^k)),
  \label{eq:traning loss of S}   
\end{equation}

where $\lambda$ is the hyper-parameter. 
By focusing on higher confidence masks $E_H^i$, we can improve the learning process for both the student and TA by guiding them towards more reliable pseudo-semantic maps during updates. This approach helps to minimize the negative impact of noisy labels from lower confidence masks produced by SE on both models. Additionally, the supervision of patches not only aids the student in refining segmentation for larger objects but also enhances the recognition of smaller objects.

Additionally, we utilize the refined boundary map $B_{ref}^i$ obtained from DARv2 to supervise the student's boundary map ($B_S^i$) in the corresponding overlapping area $O_{ij}$, enhancing the student's awareness of boundaries.

\begin{equation}
\setlength{\abovedisplayskip}{3pt}
\setlength{\belowdisplayskip}{3pt}
\begin{aligned}
\mathcal{L}_{bd}^{T,S}=\sum_{k = 1}^{H\times W}\frac{|B_{ref}^i-B_S^i|}{C_o}.
\label{eq:boundary loss of T and S}
\end{aligned}
\end{equation}

Therefore, the total loss employed for the student $\textbf{F}_{S}$ comprises three components:

\begin{equation}
\setlength{\abovedisplayskip}{5pt}
\setlength{\belowdisplayskip}{5pt}
  \mathcal{L}_{student}= \mathcal{L}_{CE}^{TA,S} +  \mathcal{L}_{CE}^{T,S} + \mathcal{L}_{bd}^{T,S}.
  \label{eq:traning loss of St}   
\end{equation}

The total loss for the TA $\textbf{F}_{TA}$ is formulated as follows:

\begin{equation}
\setlength{\abovedisplayskip}{5pt}
\setlength{\belowdisplayskip}{5pt}
  \mathcal{L}_{TA}= \mathcal{L}_{CE}^{T,TA} +  \mathcal{L}_{CC} + \mathcal{L}_{bd}^{T,TA}.
  \label{eq:traning loss of TA}   
\end{equation}

\section{Experiments}

\begin{table*}[t!]
\centering
\caption{Per-class results of the SOTA panoramic image semantic segmentation methods on DensePASS test set. (P.: Param.)}
\setlength{\tabcolsep}{2pt}
\resizebox{\textwidth}{!}{
\begin{tabular}{l|c|c|ccccccccccccccccccc}
\toprule
Method & P. (M) & mIoU  & Road  & S.W. & Build. & Wall  & Fence & Pole & Tr.L. & Tr.S. & Veget. & Terr. & Sky & Person & Rider & Car & Truck & Bus & Train & M.C. & B.C. \\ \midrule
ERFNet~\cite{romera2017erfnet}&- &16.65 & 63.59 & 18.22    & 47.01   & 9.45 & 12.79 & 17.00  & 8.12  & 6.41 & 34.24 & 10.15 & 18.43 & 4.96  & 2.31  & 46.03 & 3.19  & 0.59  & 0.00  &8.30  & 5.55   \\
PASS(ERFNet)~\cite{yang2019pass}      &- & 23.66 & 67.84 & 28.75    & 59.69    & 19.96 & 29.41 & 8.26  & 4.54          & 8.07         & 64.96      & 13.75   & 33.50 & 12.87  & 3.17  & 48.26 & 2.17  & 0.82  & 0.29  & 23.76      & 19.46   \\ 
Omni-sup(ECANet)~\cite{yang2020omnisupervised}   &- & 43.02 & \underline{81.60} & 19.46    & 81.00    & 32.02 & 39.47 & 25.54 & 3.85          & 17.38        & 79.01      & 39.75   & 94.60 & 46.39  & 12.98 & 81.96 & 49.25 & 28.29 & 0.00  & 55.36      & 29.47   \\ 
P2PDA(Adversarial)~\cite{zhang2021transfer} &- & 41.99 & 70.21 & 30.24    & 78.44    & 26.72 & 28.44 & 14.02 & 11.67         & 5.79         & 68.54      & 38.20   & 85.97 & 28.14  & 0.00  & 70.36 & 60.49 & 38.90 & 77.80 & 39.85      & 24.02   \\ 
PCS~\cite{yue2021prototypical}&25.56 & 53.83 & 78.10 & 46.24 & 86.24  & 30.33 &45.78 & 34.04  & 22.74  & 13.00 &79.98 & 33.07 & 93.44 & 47.69  & 22.53  & 79.20 & 61.59  & 67.09  & 83.26  & 58.68  & 39.80   \\
Trans4PASS-T~\cite{zhang2022bending} &13.95 & 53.18 & 78.13 & 41.19    & 85.93    & 29.88 & 37.02  & 32.54 & 21.59         & 18.94        & 78.67      &\textbf{45.20}   &93.88 & 48.54  & 16.91 & 79.58 & 65.33 & 55.76 & 84.63 & 59.05      & 37.61   \\ 
Trans4PASS-S~\cite{zhang2022bending} &24.98 & 55.22 & 78.38 & 41.58    & 86.48 & 31.54 & 45.54  & 33.92 & 22.96 & 18.27 & 79.40 & 41.07 & 93.82 & 48.85  & 23.36 &81.02 &67.31 &69.53 & 86.13 & 60.85 & 39.09   \\ 
DPPASS-T~\cite{zheng2023both}  &14.0 &55.30 &78.74 &46.29 &87.47 &\underline{48.62} &40.47 &\underline{35.38} &24.97 &17.39 &79.23 &40.85 &93.49 &52.09 &29.40 &79.19 &58.73 &47.24 &86.48 &66.60 &38.11 \\ 
DPPASS-S~\cite{zheng2023both} &25.4 &56.28 &78.99 &48.14 &87.63 &42.12 &44.85 &34.95 &27.38 &19.21 &78.55 &\underline{43.08} &92.83 &\underline{55.99} &29.10 &80.95 &61.42 &55.68 &79.70 &70.42 &38.40         \\ 
DATR-M~\cite{zheng2023look}  &4.64 &52.90 &78.71 &48.43 &86.92 &34.92 &43.90 &33.43 &22.39 &17.15 &78.55 &28.38 &93.72 &52.08 &13.24 &77.92 &56.73 &59.53 &93.98 &51.52 &34.06 \\ 
DATR-T~\cite{zheng2023look} &14.72 &54.60 &79.43 &49.70 &87.39 &37.91 &44.85 &35.06 &25.16 &19.33 &78.73 &25.75 &93.60 &53.52 &20.20 &78.07 &60.43 &55.82 &91.11 &67.03 &34.32         \\ 
DATR-S~\cite{zheng2023look} &25.76 &56.81 &80.63 &51.77 &87.80 &44.94 &43.73 &\textbf{37.23} &25.66 &\underline{21.00} &78.61 &26.68 &93.77 &54.62 &\underline{29.50} &80.03 &67.35 &63.75 &87.67 &67.57 &37.10 \\ 
GoodSAM-M(ours) &3.7 &55.93 &79.57 &51.04 &86.24 &43.42 &44.86 &30.92 &26.60 &20.62 &77.79 &25.43 &92.99 &53.77 &25.84 &82.01 &70.94 &62.29 &91.93 &58.24 &38.25\\ 
GoodSAM-T(ours) &14.0 &58.21 &80.06 &\underline{53.29} &89.75 &44.91 &46.98 &31.13 &27.81 &19.83 &79.58 &25.72 &93.81 &55.44 &26.99 &84.54 &73.07 &68.41 &93.99 &67.36 &43.39\\ 
GoodSAM-S(ours) &25.4 &\underline{60.56} &80.98 &52.96 &\textbf{93.22} &48.17 &\textbf{51.28} &33.51 &\underline{28.09} &20.15 &\underline{81.64} &30.97 &95.21 &55.13 &29.01 &87.89 &\underline{75.28} & 69.37 &\underline{94.98} &\textbf{73.28} &\underline{49.64}\\ \midrule
GoodSAM++-M (ours) & 3.7 & 56.31 & 80.47 & 52.99 & 86.05 & 44.67 & 44.07 & 32.58 & 25.00 & 20.34 & 78.40 & 25.95 & 93.17 & 53.81 & 27.21 & 81.06 & 69.19 & 64.29 & 91.55 & 60.24 & 38.19 \\
GoodSAM++-T (ours) & 14.0 & 58.98 & 79.51 & \textbf{55.08} & 91.36 & 46.63 & 48.77 & 28.92 & 25.60 & \textbf{21.62} & 77.37 & 24.45 & \underline{95.60} & 55.78 & 28.78 & \underline{86.33} & 74.86 & \underline{70.20} & \textbf{95.78} & 69.15 & 44.83 \\
GoodSAM++-S (ours) & 25.4 &\textbf{61.20} &\textbf{83.27} &51.25 &\underline{91.51} &\textbf{50.46} &\underline{49.67} &35.28 &\textbf{28.11} &18.44 &\textbf{83.93} &29.26 &\textbf{97.50} &\textbf{57.42} &\textbf{31.30} &\textbf{90.18} &\textbf{77.57} &\textbf{70.44} &93.27 &\underline{72.01} &\textbf{51.93} \\
\bottomrule
 \end{tabular}}
\label{perclass}
\end{table*}

\subsection{Datasets and Implementation Details}

\textbf{Dataset.} We leverage two outdoor benchmark datasets WildPASS~\cite{yang2021context} and DensePASS~\cite{ma2021densepass} to assess the segmentation performance of the GoodSAM++. 
The resolution of images in both datasets utilized is 400$\times$2048. 
For indoor scene comparisons, we use the Stanford2D3D Panoramic (SPan) dataset~\cite{armeni2017joint}. The resolution of the images utilized in the SPan dataset is 2048$\times$4096. Additionally, we search online for open world indoor and outdoor ERP images, and we also use a 360 camera~\cite{macpherson2022360} to capture some indoor and outdoor ERP images. We change the resolution of these images to 1024$\times$2048.

\noindent \textbf{Implementation details.} We train the proposed framework with PyTorch in 4 NVIDIA A6000 GPUs. 
We keep SAM frozen during our experiments and utilize it solely for providing instance masks and boundary information.
In outdoor dataset training, for the TA and student models, we opt for the fine-tuned Segformer~\cite{xie2021segformer} series, encompassing B0-B5 variants, which come in six different sizes and exhibit varying performance levels in 2D image semantic segmentation. 
We set the horizontal window size as $400\times 256$ and vertical window size as $200\times 512$. The hyper-parameter $\alpha$ is set to 0.3. The hyper-parameters of weight for reliable masks are set to 0.2.
In indoor dataset training, we choose Trans4PASS~\cite{zhang2022bending} and Trans4PASS+~\cite{zhang2024behind} as our TA, encompassing the tiny and small variants. We change the horizontal window size as $1024\times 512$ and vertical window size as $512\times 1024$.
\textit{More details can be found in the supplmat.}

\subsection{Comparisons with Existing Works}
\subsubsection{Outdoor Scene Comparisons}
We first compare our GoodSAM++ with GoodSAM and previous panoramic semantic segmentation methods, including ERFNet~\cite{romera2017erfnet}, PASS~\cite{yang2019pass}, Omni-sup~\cite{yang2020omnisupervised}, P2PDA~\cite{zhang2021transfer}, PCS~\cite{yue2021prototypical}, Trans4PASS~\cite{zhang2022bending}, DPPASS~\cite{zheng2023both}, and DATR~\cite{zheng2023look}, on the DensePASS dataset. As demonstrated in Tab.~\ref{perclass}, our models GoodSAM++-M, GoodSAM++-T, and GoodSAM++-S consistently outperform other approaches at their respective parameter settings. Specifically, our GoodSAM++-S significantly outperforms GoodSAM-S, DATR-S, DPPASS-S, and Trans4PASS-S, with IoU improvements of \textbf{0.64}\%, \textbf{4.39}\%, \textbf{4.92}\%, and \textbf{5.98}\% respectively, setting a new SOTA performance. Furthermore, GoodSAM++-M not only exceeds the performance of GoodSAM-M but also achieves a notable mIoU of \textbf{56.31}\% with a parameter count of just \textbf{3.7} million, making it comparable to DATR-S and outperforming Trans4PASS-S.
Regarding the segmentation performance for each class, our GoodSAM++-S excels in most categories, including  `road' (+\textbf{1.67}\% IoU), `sky' (+\textbf{2.29}\% IoU), and nearly all types of transportation (\textit{e.g.}, `car' with +\textbf{2.29}\% IoU).
Fig.~\ref{visFigure} provides visual comparisons of GoodSAM++-S against other methods on the Densepass evaluation set. These comparisons demonstrate that our GoodSAM++-S effectively generates distortion-aware and boundary-enhanced semantic segmentation maps, benefiting from the supervision of SAM and TA.

Tab.~\ref{wildpass} presents the experimental results on the WildPASS validation dataset. Our GoodSAM++ outperforms existing methods utilizing Segformer B1 and B2 backbones. Specifically, our GoodSAM++ exceeds the performance of GoodSAM across various backbone sizes, with improvements of \textbf{0.33}\% mIoU for the mini size, \textbf{0.7}\% mIoU for the tiny size, and \textbf{0.79}\% mIoU for the small size. Additionally, compared to previous UDA methods, our GoodSAM++-T surpasses DPPASS-T by \textbf{2.74}\% mIoU on the Segformer-B1 backbone, while our GoodSAM++-S outperforms DPPASS-S by \textbf{2.44}\% mIoU on the Segformer-B2 backbone. This demonstrates that even with an increased number of evaluation images, our GoodSAM++ consistently achieves superior performance.

\begin{table}[t!]
\centering
\setlength{\tabcolsep}{10pt}
\caption{Experimental results of the SOTA panoramic image semantic segmentation methods on WildPASS test set.}
\resizebox{\linewidth}{!}{
\begin{tabular}{ccc}
\toprule
Method                      & Backbone           & mIoU(\%) \\ \midrule
\multirow{2}*{Source domain Supervised} & Segformer-B1~\cite{xie2021segformer}     & 47.90         \\ 
                            & Segformer-B2~\cite{xie2021segformer}       &  54.11       \\ \midrule
Trans4PASS-T~\cite{zhang2022bending}                & Segformer-B1  &54.67          \\
Trans4PASS-S~\cite{zhang2022bending}                & Segformer-B2  &62.91          \\ 
DPPASS-T~\cite{zheng2023both}                & Segformer-B1       & 60.38       \\ 
DPPASS-S~\cite{zheng2023both}                  & Segformer-B2       & 63.53         \\ 
GoodSAM-M(ours)                  & Segformer-B0      & 58.65         \\
GoodSAM-T(ours)                  & Segformer-B1       & 62.42         \\
GoodSAM-S(ours)                  & Segformer-B2       & 65.18       \\
GoodSAM++-M(ours)                  & Segformer-B0      & 58.98         \\
GoodSAM++-T(ours)                  & Segformer-B1       & 63.12         \\
GoodSAM++-S(ours)                  & Segformer-B2       & \textbf{ 65.97 }    \\
 \bottomrule
\end{tabular}}
\label{wildpass}
\end{table}

\begin{figure*}[t!]
    \centering
    \includegraphics[width=\textwidth]{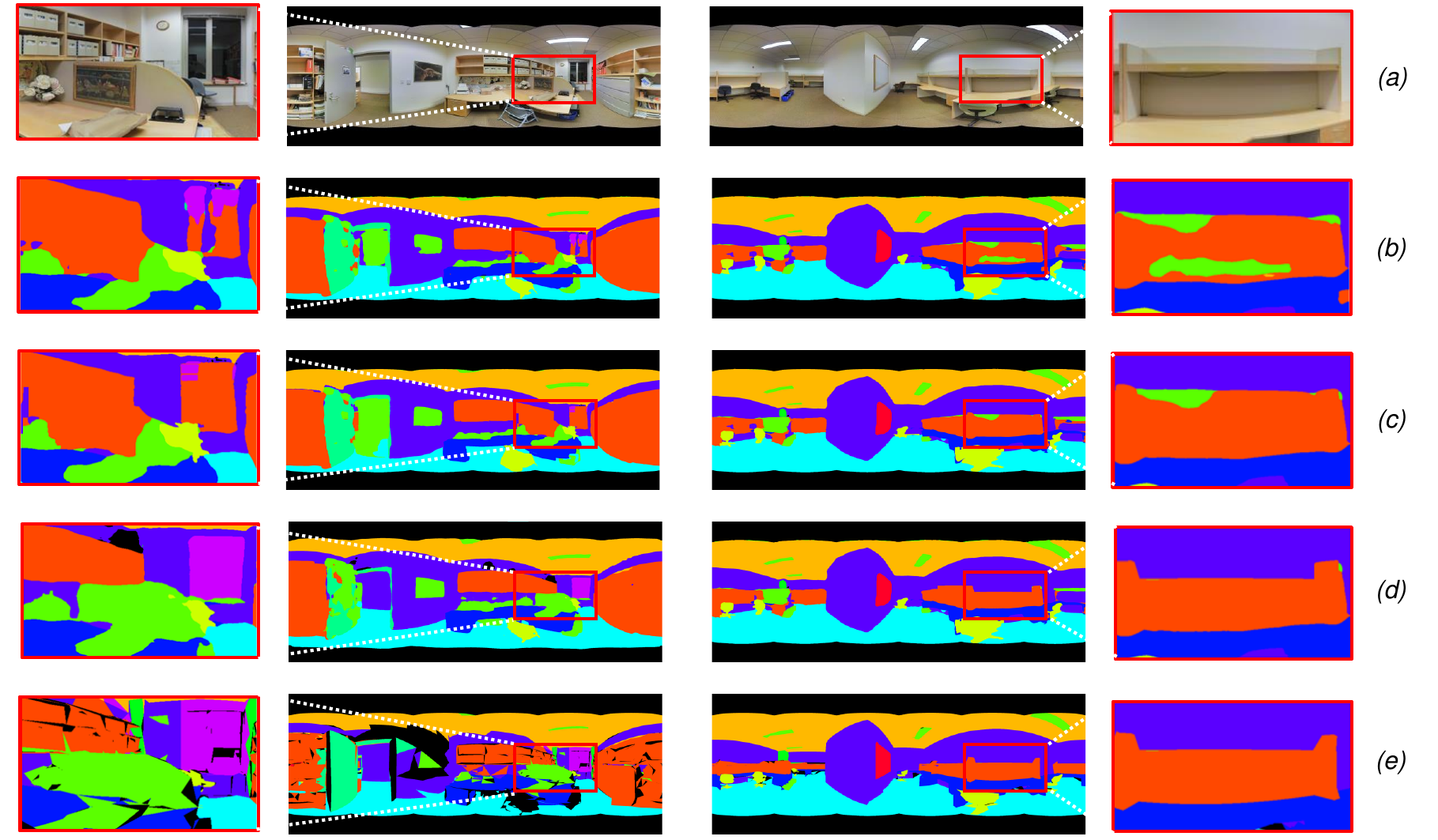}
    \caption{Example visualization results from the Stanford2D3D panoramic test set: (a) Input panorama image, (b) Trans4PASS+-S~\cite{zhang2024behind}, (c) GoodSAM-S, (d) GoodSAM++-S, (e) Ground truth.}
    \label{visFigure_indoor}
\end{figure*}

\subsubsection{Indoor Scene Comparisons}
To demonstrate the performance of our GoodSAM++ in indoor scenes, we compare it with Trans4PASS+ and the previous version of GoodSAM on the Stanford2D3D panorama evaluation dataset, as shown in \ref{visFigure_indoor}. It can be observed that our GoodSAM++-S achieves more accurate boundary segmentation performance compared to Trans4PASS+-S and GoodSAM-S. Additionally, the noise within the segmented mask regions is significantly reduced in GoodSAM++-S. This further demonstrates that our DARv2 module effectively aids our student model in obtaining distortion-aware and boundary-refined pseudo-semantic maps.

\begin{table}[t]
\centering
\caption{Ablation of different combination of sliding window strategy and CTCF block during the training process.
P.S.M represents pseudo semantic maps. }
\setlength{\tabcolsep}{8pt}
\resizebox{\linewidth}{!}{
\begin{tabular}{cccc|ccc}
\midrule
\multicolumn{4}{c}{SAM and TA}& \multicolumn{3}{c}{mIoU}\\
\midrule
SW & SWv2 & CTCF & CTCFv2  & P.S.M & TA & Student\\
\midrule
- & -& -& - &- &27.62 &15.88\\
$\surd$ & -& -& -&- &53.86 &15.88\\
\rowcolor{gray! 30}
- & $\surd$& -& -&- &53.29 &15.88\\
$\surd$ & - &$\surd$ &- &55.88 &53.86 &15.88\\
\rowcolor{yellow! 30}
- & $\surd$ &$\surd$ &- &55.35 &53.29 &15.88\\
\rowcolor{gray! 30}
- & $\surd$ &- &$\surd$ &56.17 &53.29 &15.88\\
\bottomrule
\end{tabular}}
\label{tab:Learning Scheme in Training and Testing_1}
\end{table}

\begin{table}[t]
\centering
\caption{Ablation of loss functions in DAR and DARv2 during the training process.
S.C. denotes the combination of SW and CTCF, while S.C.v2 denotes combination of SWv2 and CTCFv2. }
\setlength{\tabcolsep}{1.5pt}
\resizebox{\linewidth}{!}{
\begin{tabular}{cc|ccccc|ccc}
\midrule
\multicolumn{2}{c}{SAM and TA}& \multicolumn{5}{c}{TA} & \multicolumn{3}{c}{mIoU}\\
\midrule
S.C. & S.C.v2 & $\mathcal{L}_{ce}^{T,TA}$ & $\mathcal{L}_{CC}$ & $\mathcal{L}_{CCv2}$ & $\mathcal{L}_{bd}^{T,TA}$ & $\mathcal{L}_{bdv2}^{T,TA}$ & P.S.M & TA & Student\\
\midrule
$\surd$ & -& $\surd$& - & -& - & -  &60.68 &58.89 &49.88\\
\rowcolor{gray! 30}
- & $\surd$  & $\surd$& - & -& - & -  &60.97 &59.10 &49.93\\
$\surd$ & - & $\surd$ & $\surd$ & -& -  &- &- &60.07 &50.07\\
\rowcolor{gray! 30}
- & $\surd$  & $\surd$& - & $\surd$& - & -  &- &61.28 &50.39\\
$\surd$ & -  & $\surd$& $\surd$ & -& $\surd$& -  &- &62.49 &50.90\\
\rowcolor{yellow! 30}
$\surd$ & - & $\surd$& $\surd$ & -& -& $\surd$&- &63.17 &51.25\\
\rowcolor{gray! 30}
- & $\surd$ & $\surd$& - & $\surd$& - &$\surd$ &- &63.61 &51.40\\
\bottomrule
\end{tabular}}
\label{tab:Learning Scheme in Training and Testing_2}
\end{table}

\begin{figure}[t]
    \centering
    \includegraphics[width=0.95\columnwidth]{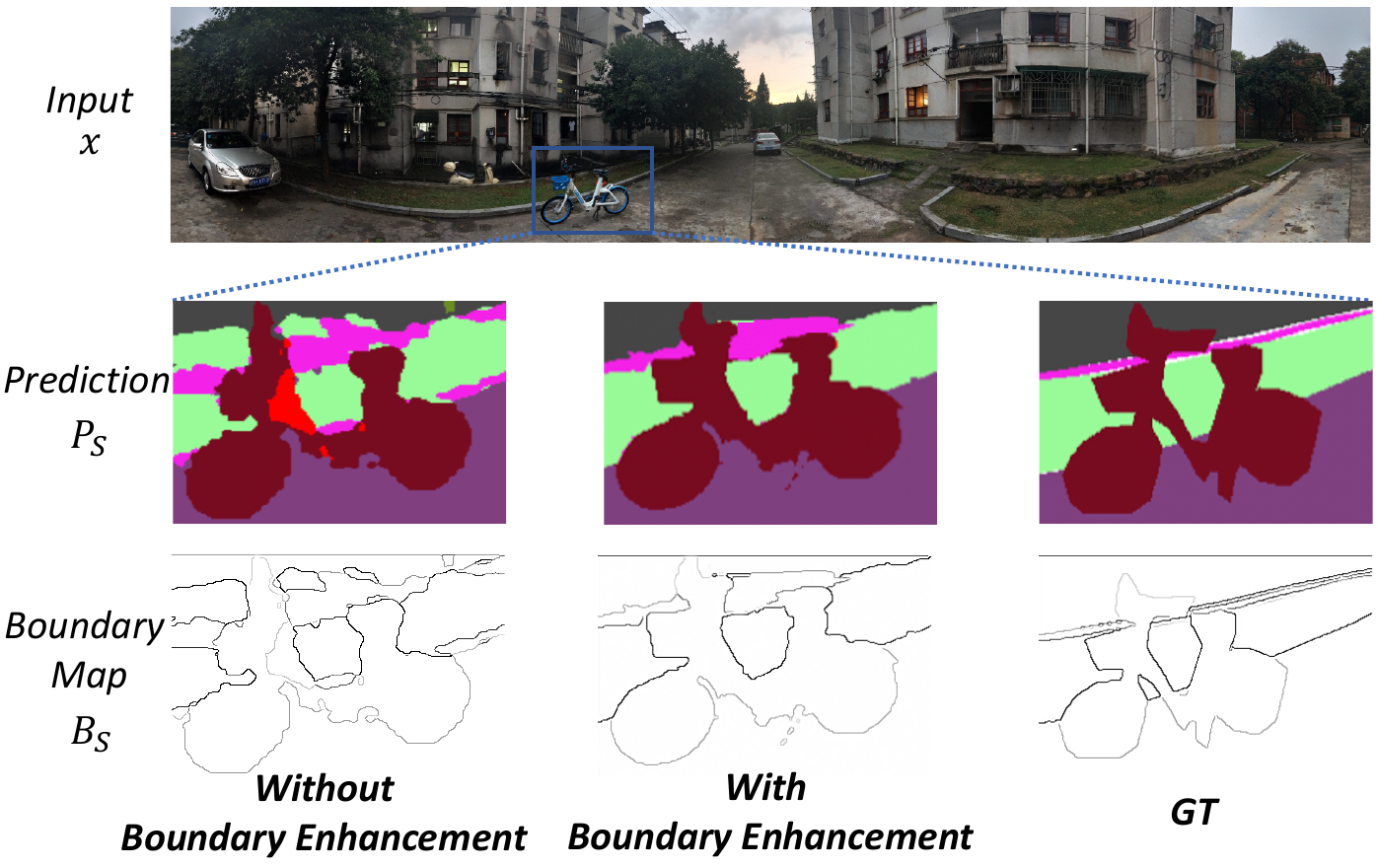}
    \caption{Effectiveness of the boundary enhancement block.
    }
    \label{fig:boundary}
\end{figure}

\begin{table}[t]
\centering
\caption{Ablation of the loss functions on our student during the training process.
P.S.M represents pseudo semantic maps. }
\setlength{\tabcolsep}{4.5pt}
\resizebox{\linewidth}{!}{
\begin{tabular}{cc|cccc|ccc}
\midrule
\multicolumn{2}{c}{SAM and TA} & \multicolumn{4}{c}{Student}& \multicolumn{3}{c}{mIoU}\\
\midrule
DARv1 & DARv2 & $\mathcal{L}_{CE}^{TA,S}$ &$\mathcal{L}_{CE}^{T,S}$ & $\mathcal{L}_{bd}^{T,S}$ & $\mathcal{L}_{bdv2}^{T,S}$ & P.S.M & TA & Student\\
\midrule
$\surd$ & - &$\surd$&-&- & - &60.68 &58.89 &49.88\\
\rowcolor{gray! 30}
- & $\surd$ &$\surd$&-&- & - &60.97 &59.10 &49.93\\
$\surd$ & - &$\surd$&$\surd$&- & - &- &62.49 &54.12\\
\rowcolor{gray! 30}
- & $\surd$ &$\surd$&$\surd$&- & - &- &63.61 &54.72\\
$\surd$ & - &$\surd$&$\surd$&$\surd$ & - &- &62.49 &55.93\\
\rowcolor{yellow! 30}
$\surd$ & - &$\surd$&$\surd$&- & $\surd$ &- &62.49 &55.98\\
\rowcolor{yellow! 30}
- & $\surd$ & $\surd$ &$\surd$&$\surd$ & - &- &63.61 &56.09\\
\rowcolor{gray! 30}
- & $\surd$ &$\surd$&$\surd$&- & $\surd$ &- &63.61 &56.31\\
\bottomrule
\end{tabular}}
\label{tab:Learning Scheme in Training and Testing_3}
\end{table}

\subsection{Ablation Studies and Analysis}
\subsubsection{Effectiveness of DAR module}
Tab.~\ref{tab:Learning Scheme in Training and Testing_1},~\ref{tab:Learning Scheme in Training and Testing_2}, and~\ref{tab:Learning Scheme in Training and Testing_3} illustrate the effectiveness of each component of our GoodSAM++ and previous GoodSAM version. We choose Segformer-B5 as the TA model and Segformer-B0 as the student model. Specifically, rows without shading represent ablation studies in GoodSAM, while rows with gray shading indicate ablation studies in GoodSAM++. Rows with yellow shading represent results that combine blocks from both GoodSAM and GoodSAM++ for further comparisons.

\noindent \textbf{1) Effectiveness of sliding window sampling.} The first three rows of Tab.~\ref{tab:Learning Scheme in Training and Testing_1} showcase a performance comparison before and after implementing sliding window sampling, highlighting a significant performance gap. This decline is attributed to Segformer being trained on 2D images; when directly applied to ERP images with larger FoV characteristics, the model exhibits a significant drop in performance. Specifically, employing the original version of the sliding window strategy in GoodSAM resulted in a performance gap that is nearly twice as significant compared to not using the sliding window strategy (TA: \textbf{27.62}\% mIoU vs. \textbf{53.86}\% mIoU).
When we update to the sliding window strategy (SWv2), it can be observed that as the number of patches increases, the re-stitched prediction map exhibits inconsistencies at the seams, leading to a decline in TA performance (B5: \textbf{53.86}\% mIoU vs. \textbf{53.29}\% mIoU).

\begin{figure}[t]
    \centering
    \includegraphics[width=\columnwidth]{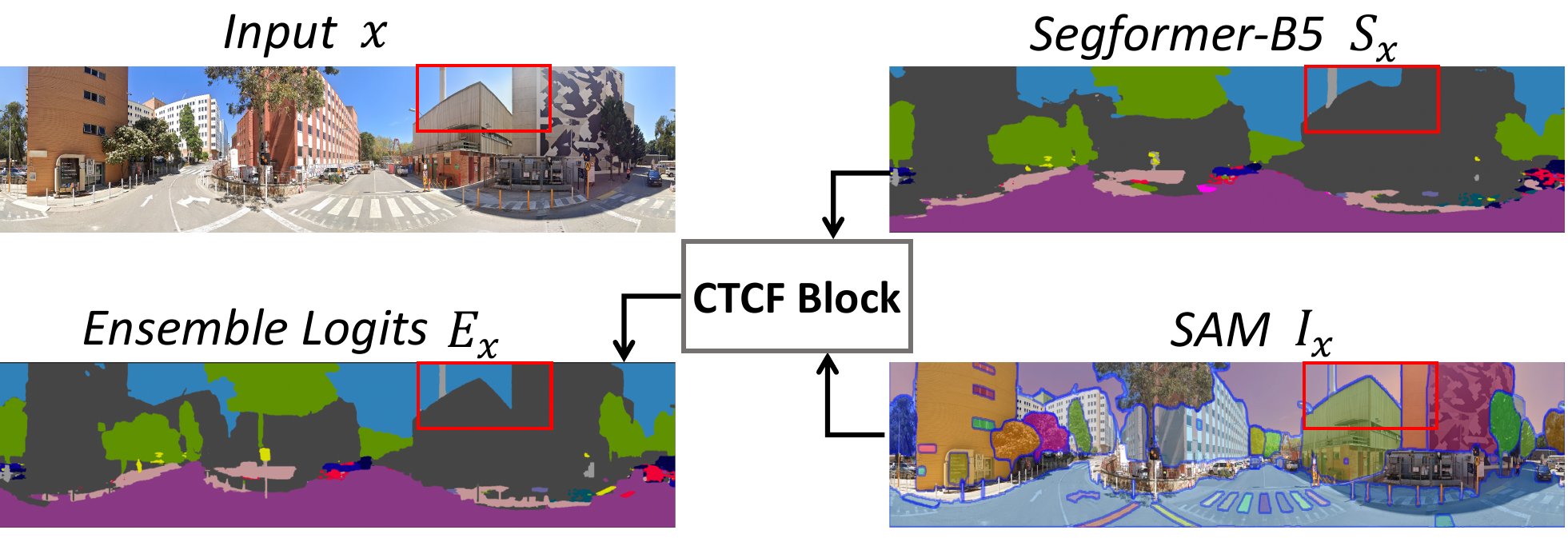}
    \caption{Effectiveness of the CTCF block.
    }
    \label{fig:CTCF}
\end{figure}

\noindent \textbf{2) Effectiveness of CTCF block.} We evaluate the CTCF and CTCFv2 blocks using the non-overlapping sliding window stitching strategy. Our approach involves generating patch pseudo-semantic maps with CTCF and CTCFv2, and subsequently reconstructing the entire ERP prediction map at the corresponding positions. As shown in Tab.~\ref{FusionW}, the mIoU of the pseudo-semantic maps generated by the previous version of CTCF currently stands at  \textbf{55.88}\% mIoU, surpassing TA's performance by \textbf{2.02}\% mIoU. Furthermore, as we utilize $\mathcal{L}_{ce}^{T,TA}$ to update TA based on patch pseudo semantic maps, we observe a continuous improvement(+\textbf{5.1}\% mIoU) in pseudo semantic maps performance alongside the enhancement of TA's performance. Fig.~\ref{fig:CTCF} demonstrates the visual differences between the segmentation map produced by Segformer-B5 as TA and the pseudo semantic maps obtained through CTCF block,  highlighting the effectiveness of our CTCF block. Tab.~\ref{tab:Learning Scheme in Training and Testing_1} and  Tab.~\ref{tab:Learning Scheme in Training and Testing_2} also shows the results of combining SWv2 and CTCFv2. It can be observed that the performance of the pseudo semantic maps improved from \textbf{55.35}\% mIoU to \textbf{56.17}\% mIoU. This demonstrates that our CTCFv2 not only overcomes the performance decline caused by changing SW but also helps TA achieve more efficient performance improvement(B5: \textbf{58.89}\% mIoU vs. \textbf{59.10}\% mIoU).


\noindent \textbf{3) Effectiveness of consistency constraint.}
we now ablate the consistency constraint based on the overlapping sliding window. Introducing the consistency constraint $\mathcal{L}_{CC}$ to implicitly mitigate the inconsistencies in overlapping regions between adjacent windows caused by distortion significantly improves the TA's performance(+\textbf{1.18}\% mIoU).
When we use $\mathcal{L}_{CCv2}$ with SWv2, the performance of TA further improves (+\textbf{2.18}\% mIoU) compared to the combination of original SW and $\mathcal{L}_{CC}$. This demonstrates that CCv2 can more efficiently help our TA adapt to panoramic images by constraining the prediction-level consistency in overlapping regions with significant differences at the prediction level.

\begin{figure}[t]
    \centering
    \includegraphics[width=0.95\columnwidth]{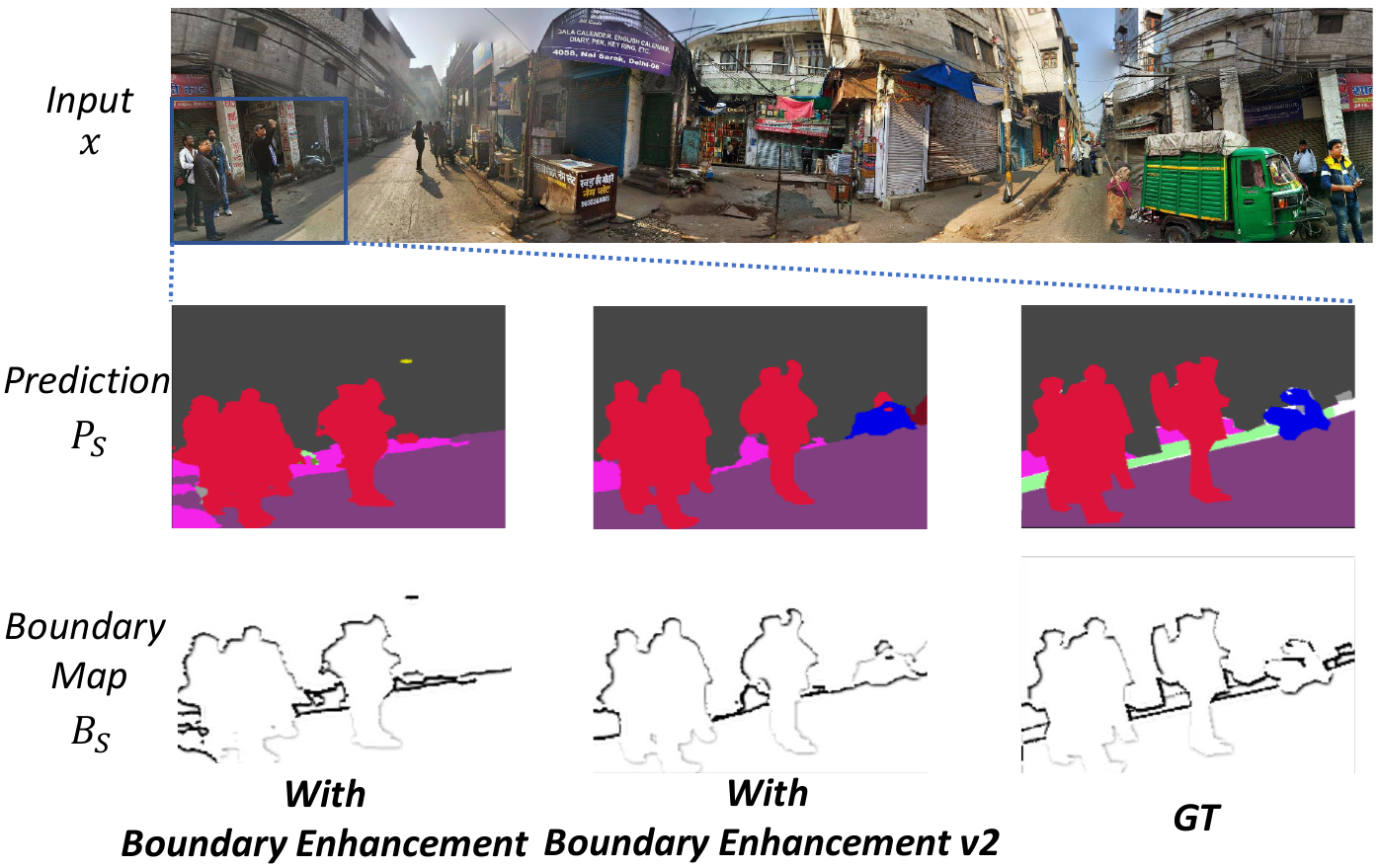}
    \caption{Visual comparison of the BEv2 and original BE block.
    }
    \label{fig:boundaryv2}
\end{figure}

\noindent \textbf{4) Effectiveness of boundary enhancement.} We propose the boundary enhancement block to utilize the boundary information provided by SAM, thereby enhancing TA's ability to accurately predict boundary pixels. As revealed by Tab.~\ref{tab:Learning Scheme in Training and Testing_2}, in GoodSAM framework, the introduced boundary-enhanced loss boosts TA's performance by \textbf{2.42}\% mIoU (from\textbf{ 60.07}\% mIoU to \textbf{62.49}\% mIoU). The right section of visualization results in Fig.~\ref{visFigure} demonstrates the superior boundary segmentation performance of our GoodSAM for 'fence' and 'sidewalk' compared to previous methods. Meanwhile, Fig.~\ref{fig:boundary} also illustrates the positive impact of the boundary enhancement block on student predictions. These experimental results indicate that our boundary enhancement block effectively aids both TA and the student model in increasing boundary awareness and addressing distortion issues. We further discuss the performance of our improved BEv2 block. The experiment in the yellow-shaded row in Tab.~\ref{tab:Learning Scheme in Training and Testing_2} shows the results when we replace the BE block with our proposed BEv2 block while finetuning TA. It can be observed that TA's performance improved from \textbf{62.49}\% mIoU to \textbf{63.17}\% mIoU. This indicates that the BEv2 block can obtain more reliable boundary maps in overlapping regions. Furthermore, when we combine the BEv2 block with our SWv2 and CCv2 in the GoodSAM++ framework, we find that TA's performance further improves after finetuning, reaching \textbf{63.61}\% mIoU. Meanwhile, Fig.~\ref{fig:boundaryv2} also illustrates the superior performance of BEv2 compared to the original BE Block. This demonstrates that our DARv2 module can better help TA mitigate distortion issues in ERP compared to the original DAR module.

\begin{figure*}[t!]
    \centering
    \includegraphics[width=\textwidth]{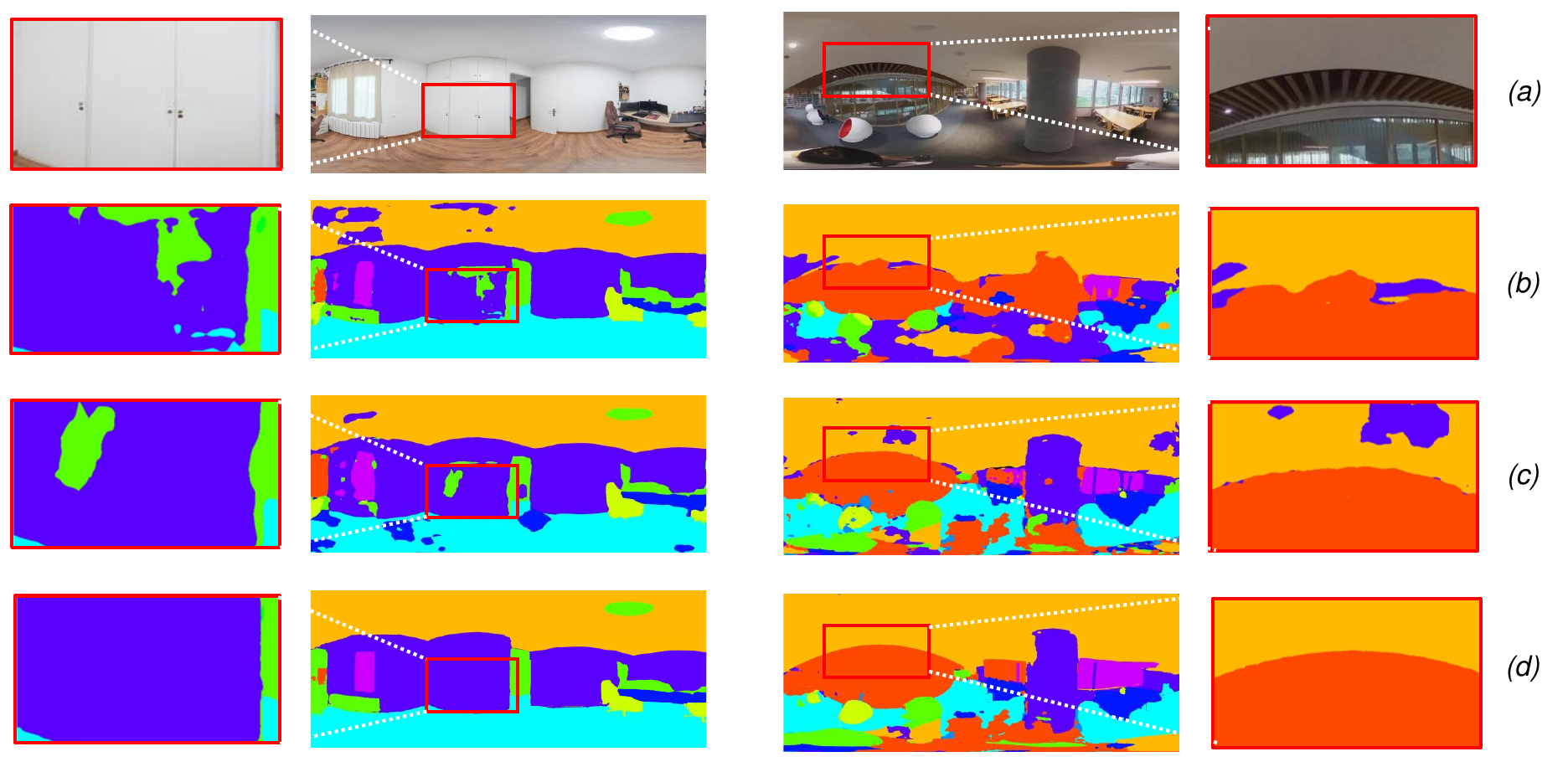}
    \caption{Example visualization results from the indoor open-world and our self-collected dataset: (a) Input panorama image, (b) Tran4PASS+-S~\cite{zhang2024behind} (c) GoodSAM-S, (d) GoodSAM++-S.}
    \label{visFigure_OPENINDOOR}
\end{figure*}

\begin{figure*}[t!]
    \centering
    \includegraphics[width=\textwidth]{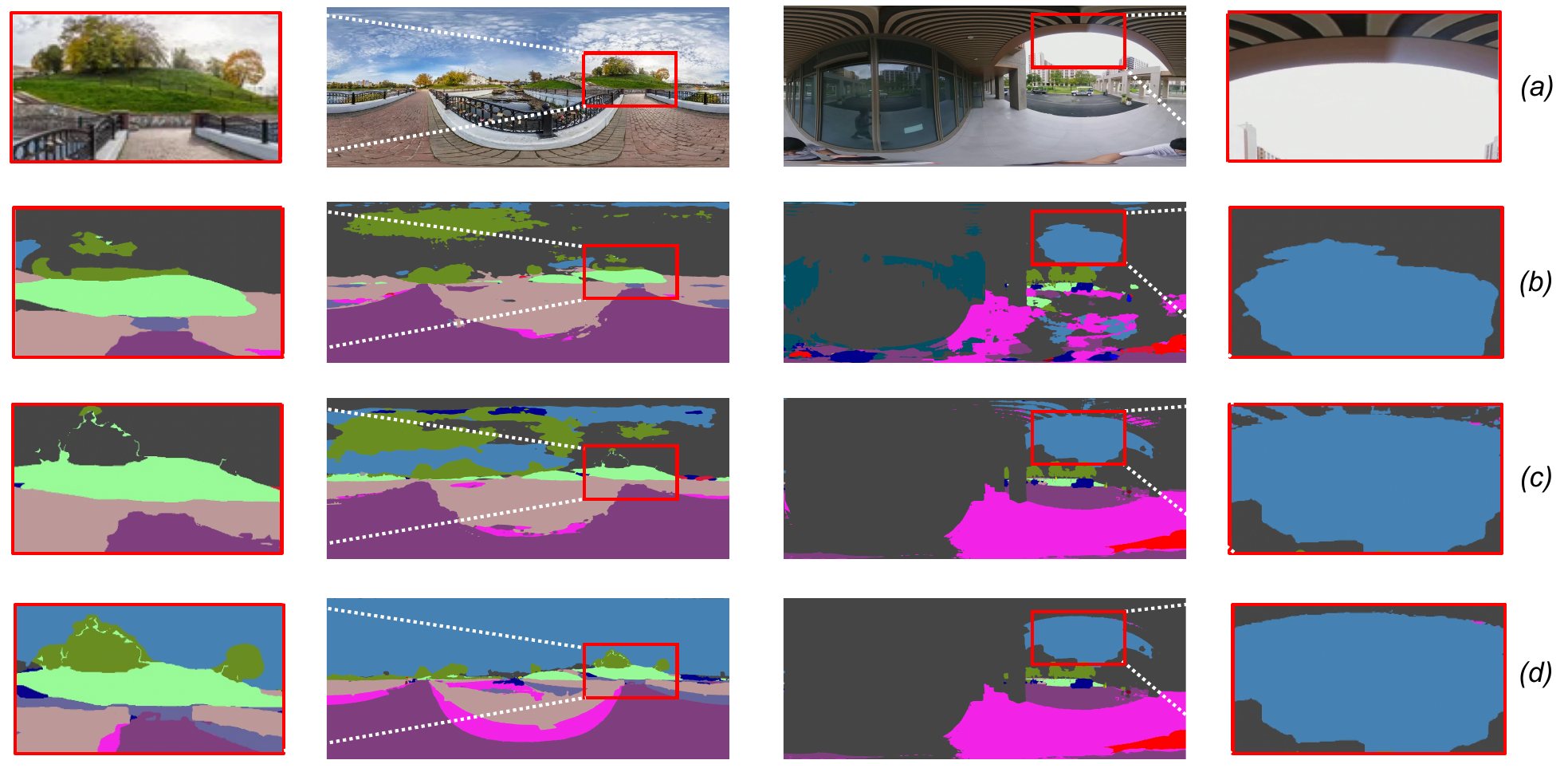}
    \caption{Example  visualization results from the outdoor open-world and our self-collected dataset: (a) Input panorama image, (b) Segformer-B5~\cite{xie2021segformer} without sliding window sampling, (c) GoodSAM-S, (d) GoodSAM++-S.}
    \label{visFigure_OPENOUTDOOR}
\end{figure*}

\begin{figure*}[t!]
    \centering
    \includegraphics[width=\textwidth]{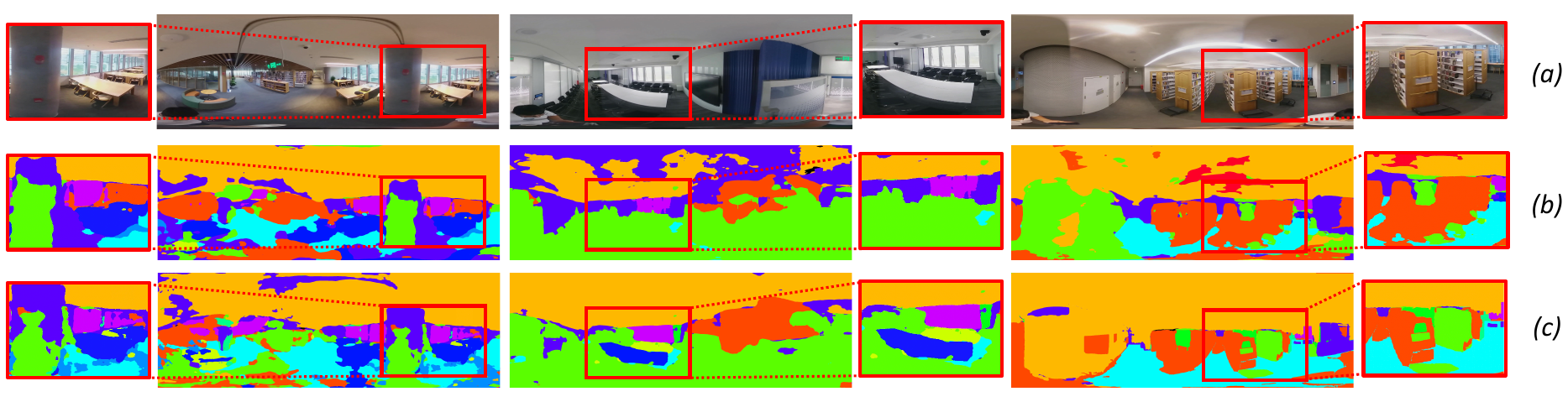}
    \caption{More visualization results from the  our self-collected indoor dataset: (a) Input panorama image, (b) GoodSAM-S, (c) GoodSAM++-S.}
    \label{visFigure_indooradd}
\end{figure*}

\subsubsection{Effectiveness of MKA module}
For the MKA module, as shown in Tab.~\ref{tab:Learning Scheme in Training and Testing_1} and~\ref{tab:Learning Scheme in Training and Testing_3}, we observe that the performance of the student is only \textbf{15.88}\% mIoU when the entire ERP image is directly input. However, as we update the student using TA's logits, the performance increases to \textbf{49.93}\% mIoU with the improvement of TA's performance. 
Furthermore, by supervising the student's semantic logits in the corresponding region using window-based pseudo semantic maps, the performance of the GoodSAM student reached \textbf{54.12}\% mIoU. Building on the DARv2 foundation, our GoodSAM++ student further improved to \textbf{54.72}\% mIoU.
When we use the refined boundary map output from the BE block to provide boundary-level supervision to our GoodSAM student model, its performance reaches \textbf{1.81}\% mIoU. Furthermore, using the output from the BEv2 block on the same previous DAR version, the performance of the GoodSAM++ student model shows further improvement over GoodSAM. This indicates that our BEv2 block can achieve more accurate boundary pixel prediction. Additionally, combining the previous version of boundary loss with the DARv2 module, we observe that as the TA's performance improves, our student model also sees an enhancement, reaching \textbf{56.09}\% mIoU. Finally, when combining the outputs of DARv2 and the new BEv2 to supervise our GoodSAM++ student model, it achieves the best performance, attaining \textbf{56.31}\% mIoU.
This indicates the  effectiveness of MKA in endowing GoodSAM and GoodSAM++ with distortion-aware and boundary-aware capabilities.

\subsubsection{Other Analysis}
\begin{table}[t]
\centering
\caption{Ablation of fusion ways and different TA model.}
\setlength{\tabcolsep}{4.8pt}
\resizebox{0.48\textwidth}{!}{
\begin{tabular}{ccccccccc}
\toprule
& & $B_0$  & $B_1$  & $B_2$  & $B_3$  & $B_4$ & $B_5$ \\ \midrule
&SSA~\cite{chen2023semantic}  &39.28 & 40.28 &47.93 & 49.13 & 52.19 & 54.32    \\ 
\multirow{2}*{Fusion}&SEPL~\cite{chen2023segment}  &39.68 &40.75 & 48.98 &50.37   &52.53 &54.58  \\ 
&CTCF &40.68 &41.75 & 49.7 &52.89   &54.13 &55.88 \\ 
\midrule
Student&$B_2$ &- &- &- &57.16 &59.02 &60.56&\\
\bottomrule
\end{tabular}}

\label{FusionW}
\end{table}

\noindent \textbf{Analysis about CTCF module.}
We introduce the CTCFv2 block to adaptively combine the outputs of SAM and TA, obtaining patch pseudo semantic maps. We evaluate our previous version of CTCF by comparing it with current methods that combine instance masks and semantic logits~\cite{chen2023semantic,chen2023segment}. SSA~\cite{chen2023semantic} assigns the label of the instance mask as the label that appears most frequently in the corresponding region of the semantic map. The fusion mechanism in SEPL~\cite{chen2023segment} is similar to SSA. They analyze each instance mask, selecting the label if it occupies more than half of the area in the corresponding region of the semantic map or if the distribution is almost covered by the instance mask.

However, when analyzing instance masks of various sizes during our experiments, we find that medium-sized masks often show a comparable coverage rate between the top two labels. This increases the risk of errors if the label with the highest rate is selected directly. To address this, we set different thresholds for different mask sizes and integrate SE to determine the highest-confidence label, allowing for an adaptive fusion of instance masks and semantic logits. As shown in Tab.~\ref{FusionW}, our fusion mechanism significantly enhances the robustness and accuracy of the fusion process compared to the other two methods.
The comparative experiments highlighted by the yellow shading in Tab.~\ref{tab:Learning Scheme in Training and Testing_1} also demonstrate that, based on SWv2, the pseudo semantic map output by our enhanced CTCFv2 block shows better performance compared to the CTCF block.



\begin{figure}[t]
    \centering
    \caption{Ablation study for $\lambda$ of knowledge transfer from pseudo semantic maps to our student. 
    }
    \includegraphics[width=0.95\columnwidth]{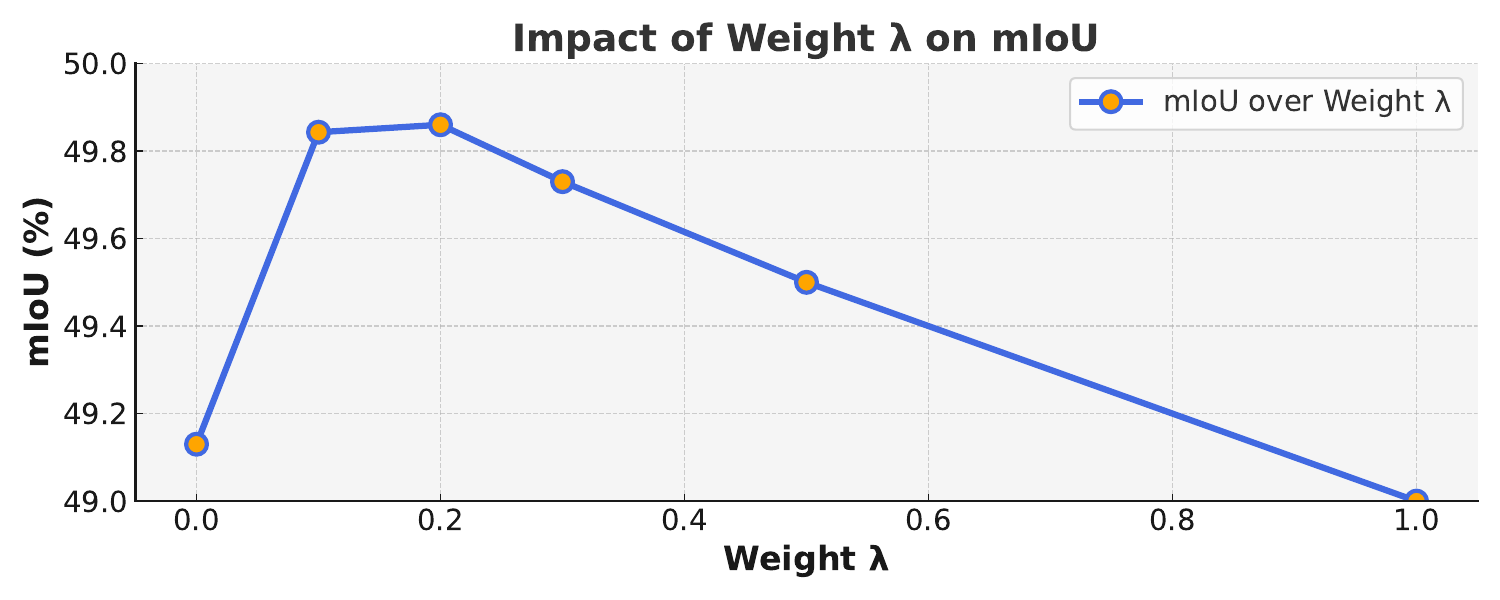}
    
    \label{fig:weight}
\end{figure}

\noindent \textbf{Hyper-parameter analysis in the loss functions.}
To transfer knowledge from window-based pseudo-semantic maps to the TA and student, we prioritize instance masks with higher confidence by assigning them greater weights for the CTCFv2 block.  Fig.~\ref{fig:weight} shows that with $\lambda$ set to 0.2, our GoodSAM++ can more efficiently extract correct knowledge from pseudo semantic maps.

\noindent \textbf{The impact of TA selection.}
In this section, we assess the impact of different TAs on the training of the student based on GoodSAM framework. As shown in Tab.~\ref{FusionW}, we observe that as the performance of TA improves, our student's performance also increases. Specifically, when TA is Segformer-B5, our student achieves a performance gain of +\textbf{3.4}\% mIoU compared to TA being Segformer-B3. This improvement is attributed to Segformer-B5's ability to provide more comprehensive and accurate semantic logits.

\noindent \textbf{The generalization of our GoodSAM++ in indoor scene.}
In this section, we compare the generalization capabilities of GoodSAM++-S and GoodSAM-S in indoor scenes. To fully demonstrate our student's performance, we also include the visualization results of Trans4PASS+-S for comparison. As shown in Fig.~\ref{visFigure_OPENINDOOR}, the left image is an indoor image in an open-world scenario found online, while the right image is a library image captured by ourselves using a 360 camera (\textit{More details can be found in the supplmat.}). It is evident that compared to Trans4PASS+-S and GoodSAM-S, our GoodSAM++-S not only achieves more accurate mask boundary predictions but also exhibits significantly less noise. This demonstrates that our GoodSAM++-S has a stronger generalization capability compared to GoodSAM-S in the open-world indoor scene.

\noindent \textbf{The generalization of our GoodSAM++ in outdoor scene.}
In this section, we compare the generalization capabilities of GoodSAM++-S and GoodSAM-S in outdoor scenes. To fully demonstrate our student's performance, we also include the visualization results of our TA model (Segformer-B5) for comparison. As shown in Fig.~\ref{visFigure_OPENOUTDOOR}, the left image is an outdoor image in an open-world scenario found online, while the right image is a library image captured by ourselves. It is evident that compared to Trans4PASS+-S and GoodSAM-S, our GoodSAM++-S also achieves more accurate mask boundary predictions but also exhibits significantly less noise in the outdoor scene. This also demonstrates that our GoodSAM++-S has a stronger generalization capability compared to GoodSAM-S in the open-world indoor scene. To further demonstrate the performance of our GoodSAM++ compared to GoodSAM in indoor scenes, Fig.~\ref{visFigure_indooradd} utilizes more of our self-collected indoor panoramic data for a visual comparison of performance. It can be observed that our GoodSAM++ achieves better segmentation results.

\noindent \textbf{The quantitative comparison of indoor scene.}
When training and testing on the Span dataset, we only used visual comparison to compare GoodSAM++ and UDA methods, \ie, Trans4PASS+, without using numerical comparisons. This is because SAM tends to produce large instance masks in indoor datasets, leading to significant label assignment errors during fusion in the CTCFv2 block. Additionally, the performance of our student model in indoor scenes is also influenced by the TA's segmentation performance in the same environments.
These factors cause our student model to learn incorrect pseudo-labels in some indoor scenarios (\ie, especially in scenes with large glass areas), thereby reducing its performance. Since previous indoor semantic segmentation methods have relied on labeled data rather than operating in a purely unsupervised manner, they have ground truth to supervise extreme classes. However, our pseudo-semantic map is constrained by the TA's segmentation maps and SAM's outputs. Moreover, we only used the target domain from the UDA method to split the training and evaluation sets, which makes the numerical comparisons unfair.
Therefore, when comparing performance in indoor scenarios, we only consider visual comparison.

\section{Conclusion and Future Work}
In this paper, we designed a more comprehensive framework, \ie, GoodSAM++ based on GoodSAM, for lightweight panorama semantic segmentation, which leverages the assistance of SAM and TA. 
We further optimized the original DAR module to enhance the TA's ability to address distortion and large Field of View (FoV) problems in panoramic images, as well as to produce distortion-aware and boundary-enhanced logits. Experimental results demonstrated that our GoodSAM++ surpasses previous SOTA UDA methods and GoodSAM across various model parameter levels.

\noindent \textbf{Limitation:} 
Although our GoodSAM++ can generate distortion-aware pseudo semantic maps, during training we found that SAM's "segment everything" mode in indoor scenes often results in large-scale prediction errors of independent instance masks. This leads to incorrect label assignments in our CTCFv2 block, which in turn degrades the performance of our student model. Furthermore, although our GoodSAM++ demonstrated superior generalization capability compared to GoodSAM in the experiments, due to the small scale of training data and fewer model parameters, GoodSAM++ struggles to achieve the zero-shot capabilities of foundational models like SAM.

\noindent \textbf{Future work:} 
It would be beneficial to fine-tune SAM to develop a foundational segmentation model tailored for panoramic images. Additionally, we aim to explore methods for distilling SAM's zero-shot capabilities into our compact segmentation model.


\bibliographystyle{IEEEtran}
\bibliography{main}

\begin{thebibliography}{10}
\providecommand{\url}[1]{#1}
\csname url@samestyle\endcsname
\providecommand{\newblock}{\relax}
\providecommand{\bibinfo}[2]{#2}
\providecommand{\BIBentrySTDinterwordspacing}{\spaceskip=0pt\relax}
\providecommand{\BIBentryALTinterwordstretchfactor}{4}
\providecommand{\BIBentryALTinterwordspacing}{\spaceskip=\fontdimen2\font plus
\BIBentryALTinterwordstretchfactor\fontdimen3\font minus \fontdimen4\font\relax}
\providecommand{\BIBforeignlanguage}[2]{{%
\expandafter\ifx\csname l@#1\endcsname\relax
\typeout{** WARNING: IEEEtran.bst: No hyphenation pattern has been}%
\typeout{** loaded for the language `#1'. Using the pattern for}%
\typeout{** the default language instead.}%
\else
\language=\csname l@#1\endcsname
\fi
#2}}
\providecommand{\BIBdecl}{\relax}
\BIBdecl

\bibitem{zheng2023look}
X.~Zheng, T.~Pan, Y.~Luo, and L.~Wang, ``Look at the neighbor: Distortion-aware unsupervised domain adaptation for panoramic semantic segmentation,'' in \emph{Proceedings of the IEEE/CVF International Conference on Computer Vision}, 2023, pp. 18\,687--18\,698.

\bibitem{wang2018self}
F.-E. Wang, H.-N. Hu, H.-T. Cheng, J.-T. Lin, S.-T. Yang, M.-L. Shih, H.-K. Chu, and M.~Sun, ``Self-supervised learning of depth and camera motion from 360 $\{$$\backslash$deg$\}$ videos,'' \emph{arXiv preprint arXiv:1811.05304}, 2018.

\bibitem{wang2020bifuse}
F.-E. Wang, Y.-H. Yeh, M.~Sun, W.-C. Chiu, and Y.-H. Tsai, ``Bifuse: Monocular 360 depth estimation via bi-projection fusion,'' in \emph{Proceedings of the IEEE/CVF Conference on Computer Vision and Pattern Recognition}, 2020, pp. 462--471.

\bibitem{jayasuriya2020active}
M.~Jayasuriya, R.~Ranasinghe, and G.~Dissanayake, ``Active perception for outdoor localisation with an omnidirectional camera,'' in \emph{2020 IEEE/RSJ International Conference on Intelligent Robots and Systems (IROS)}.\hskip 1em plus 0.5em minus 0.4em\relax IEEE, 2020, pp. 4567--4574.

\bibitem{xu2018predicting}
M.~Xu, Y.~Song, J.~Wang, M.~Qiao, L.~Huo, and Z.~Wang, ``Predicting head movement in panoramic video: A deep reinforcement learning approach,'' \emph{IEEE transactions on pattern analysis and machine intelligence}, vol.~41, no.~11, pp. 2693--2708, 2018.

\bibitem{ai2022deep}
H.~Ai, Z.~Cao, J.~Zhu, H.~Bai, Y.~Chen, and L.~Wang, ``Deep learning for omnidirectional vision: A survey and new perspectives,'' \emph{arXiv preprint arXiv:2205.10468}, 2022.

\bibitem{zheng2023both}
X.~Zheng, J.~Zhu, Y.~Liu, Z.~Cao, C.~Fu, and L.~Wang, ``Both style and distortion matter: Dual-path unsupervised domain adaptation for panoramic semantic segmentation,'' in \emph{Proceedings of the IEEE/CVF Conference on Computer Vision and Pattern Recognition}, 2023, pp. 1285--1295.

\bibitem{li2023sgat4pass}
X.~Li, T.~Wu, Z.~Qi, G.~Wang, Y.~Shan, and X.~Li, ``Sgat4pass: Spherical geometry-aware transformer for panoramic semantic segmentation,'' \emph{arXiv preprint arXiv:2306.03403}, 2023.

\bibitem{zhang2021transfer}
J.~Zhang, C.~Ma, K.~Yang, A.~Roitberg, K.~Peng, and R.~Stiefelhagen, ``Transfer beyond the field of view: Dense panoramic semantic segmentation via unsupervised domain adaptation,'' \emph{IEEE Transactions on Intelligent Transportation Systems}, vol.~23, no.~7, pp. 9478--9491, 2021.

\bibitem{zhang2022bending}
J.~Zhang, K.~Yang, C.~Ma, S.~Rei{\ss}, K.~Peng, and R.~Stiefelhagen, ``Bending reality: Distortion-aware transformers for adapting to panoramic semantic segmentation,'' in \emph{Proceedings of the IEEE/CVF conference on computer vision and pattern recognition}, 2022, pp. 16\,917--16\,927.

\bibitem{xie2021segformer}
E.~Xie, W.~Wang, Z.~Yu, A.~Anandkumar, J.~M. Alvarez, and P.~Luo, ``Segformer: Simple and efficient design for semantic segmentation with transformers,'' \emph{Advances in Neural Information Processing Systems}, vol.~34, pp. 12\,077--12\,090, 2021.

\bibitem{yang2019pass}
K.~Yang, X.~Hu, L.~M. Bergasa, E.~Romera, and K.~Wang, ``Pass: Panoramic annular semantic segmentation,'' \emph{IEEE Transactions on Intelligent Transportation Systems}, vol.~21, no.~10, pp. 4171--4185, 2019.

\bibitem{yang2020omnisupervised}
K.~Yang, X.~Hu, Y.~Fang, K.~Wang, and R.~Stiefelhagen, ``Omnisupervised omnidirectional semantic segmentation,'' \emph{IEEE Transactions on Intelligent Transportation Systems}, vol.~23, no.~2, pp. 1184--1199, 2020.

\bibitem{liu2021pano}
M.~Liu, S.~Wang, Y.~Guo, Y.~He, and H.~Xue, ``Pano-sfmlearner: Self-supervised multi-task learning of depth and semantics in panoramic videos,'' \emph{IEEE Signal Processing Letters}, vol.~28, pp. 832--836, 2021.

\bibitem{zhang2021deeppanocontext}
C.~Zhang, Z.~Cui, C.~Chen, S.~Liu, B.~Zeng, H.~Bao, and Y.~Zhang, ``Deeppanocontext: Panoramic 3d scene understanding with holistic scene context graph and relation-based optimization,'' in \emph{Proceedings of the IEEE/CVF International Conference on Computer Vision}, 2021, pp. 12\,632--12\,641.

\bibitem{ma2021densepass}
C.~Ma, J.~Zhang, K.~Yang, A.~Roitberg, and R.~Stiefelhagen, ``Densepass: Dense panoramic semantic segmentation via unsupervised domain adaptation with attention-augmented context exchange,'' in \emph{2021 IEEE International Intelligent Transportation Systems Conference (ITSC)}.\hskip 1em plus 0.5em minus 0.4em\relax IEEE, 2021, pp. 2766--2772.

\bibitem{zhang2022behind}
J.~Zhang, K.~Yang, H.~Shi, S.~Rei{\ss}, K.~Peng, C.~Ma, H.~Fu, P.~H. Torr, K.~Wang, and R.~Stiefelhagen, ``Behind every domain there is a shift: Adapting distortion-aware vision transformers for panoramic semantic segmentation,'' \emph{arXiv preprint arXiv:2207.11860}, 2022.

\bibitem{yang2020ds}
K.~Yang, X.~Hu, H.~Chen, K.~Xiang, K.~Wang, and R.~Stiefelhagen, ``Ds-pass: Detail-sensitive panoramic annular semantic segmentation through swaftnet for surrounding sensing,'' in \emph{2020 IEEE Intelligent Vehicles Symposium (IV)}.\hskip 1em plus 0.5em minus 0.4em\relax IEEE, 2020, pp. 457--464.

\bibitem{kirillov2023segment}
A.~Kirillov, E.~Mintun, N.~Ravi, H.~Mao, C.~Rolland, L.~Gustafson, T.~Xiao, S.~Whitehead, A.~C. Berg, W.-Y. Lo \emph{et~al.}, ``Segment anything,'' \emph{arXiv preprint arXiv:2304.02643}, 2023.

\bibitem{singh2022flava}
A.~Singh, R.~Hu, V.~Goswami, G.~Couairon, W.~Galuba, M.~Rohrbach, and D.~Kiela, ``Flava: A foundational language and vision alignment model,'' in \emph{Proceedings of the IEEE/CVF Conference on Computer Vision and Pattern Recognition}, 2022, pp. 15\,638--15\,650.

\bibitem{nguyen2023lvm}
D.~M. Nguyen, H.~Nguyen, N.~T. Diep, T.~N. Pham, T.~Cao, B.~T. Nguyen, P.~Swoboda, N.~Ho, S.~Albarqouni, P.~Xie \emph{et~al.}, ``Lvm-med: Learning large-scale self-supervised vision models for medical imaging via second-order graph matching,'' \emph{arXiv preprint arXiv:2306.11925}, 2023.

\bibitem{wu2023medical}
J.~Wu, R.~Fu, H.~Fang, Y.~Liu, Z.~Wang, Y.~Xu, Y.~Jin, and T.~Arbel, ``Medical sam adapter: Adapting segment anything model for medical image segmentation,'' \emph{arXiv preprint arXiv:2304.12620}, 2023.

\bibitem{ma2023segment}
J.~Ma and B.~Wang, ``Segment anything in medical images,'' \emph{arXiv preprint arXiv:2304.12306}, 2023.

\bibitem{zhang2023segment}
Y.~Zhang and R.~Jiao, ``How segment anything model (sam) boost medical image segmentation?'' \emph{arXiv preprint arXiv:2305.03678}, 2023.

\bibitem{zhang2024goodsam}
W.~Zhang, Y.~Liu, X.~Zheng, and L.~Wang, ``Goodsam: Bridging domain and capacity gaps via segment anything model for distortion-aware panoramic semantic segmentation,'' 2024.

\bibitem{zhang2024behind}
J.~Zhang, K.~Yang, H.~Shi, S.~Rei{\ss}, K.~Peng, C.~Ma, H.~Fu, P.~H. Torr, K.~Wang, and R.~Stiefelhagen, ``Behind every domain there is a shift: Adapting distortion-aware vision transformers for panoramic semantic segmentation,'' \emph{IEEE Transactions on Pattern Analysis and Machine Intelligence}, 2024.

\bibitem{orhan2022semantic}
S.~Orhan and Y.~Bastanlar, ``Semantic segmentation of outdoor panoramic images,'' \emph{Signal, Image and Video Processing}, vol.~16, no.~3, pp. 643--650, 2022.

\bibitem{yang2019can}
K.~Yang, X.~Hu, L.~M. Bergasa, E.~Romera, X.~Huang, D.~Sun, and K.~Wang, ``Can we pass beyond the field of view? panoramic annular semantic segmentation for real-world surrounding perception,'' in \emph{2019 IEEE Intelligent Vehicles Symposium (IV)}.\hskip 1em plus 0.5em minus 0.4em\relax IEEE, 2019, pp. 446--453.

\bibitem{xu2019semantic}
Y.~Xu, K.~Wang, K.~Yang, D.~Sun, and J.~Fu, ``Semantic segmentation of panoramic images using a synthetic dataset,'' in \emph{Artificial Intelligence and Machine Learning in Defense Applications}, vol. 11169.\hskip 1em plus 0.5em minus 0.4em\relax SPIE, 2019, pp. 90--104.

\bibitem{zheng2024semantics}
X.~Zheng, P.~Zhou, A.~Vasilakos, and L.~Wang, ``Semantics, distortion, and style matter: Towards source-free uda for panoramic segmentation,'' 2024.

\bibitem{zheng2024360sfuda++}
X.~Zheng, P.~Zhou, A.~V. Vasilakos, and L.~Wang, ``360sfuda++: Towards source-free uda for panoramic segmentation by learning reliable category prototypes,'' \emph{arXiv preprint arXiv:2404.16501}, 2024.

\bibitem{guttikonda2024single}
S.~Guttikonda and J.~Rambach, ``Single frame semantic segmentation using multi-modal spherical images,'' in \emph{Proceedings of the IEEE/CVF Winter Conference on Applications of Computer Vision}, 2024, pp. 3222--3231.

\bibitem{zhu2023patch}
J.~Zhu, H.~Bai, and L.~Wang, ``Patch-mix transformer for unsupervised domain adaptation: A game perspective,'' in \emph{Proceedings of the IEEE/CVF Conference on Computer Vision and Pattern Recognition}, 2023, pp. 3561--3571.

\bibitem{zhu2023good}
J.~Zhu, Y.~Luo, X.~Zheng, H.~Wang, and L.~Wang, ``A good student is cooperative and reliable: Cnn-transformer collaborative learning for semantic segmentation,'' in \emph{Proceedings of the IEEE/CVF International Conference on Computer Vision}, 2023, pp. 11\,720--11\,730.

\bibitem{zhang2017curriculum}
Y.~Zhang, P.~David, and B.~Gong, ``Curriculum domain adaptation for semantic segmentation of urban scenes,'' in \emph{Proceedings of the IEEE international conference on computer vision}, 2017, pp. 2020--2030.

\bibitem{roy2023sam}
S.~Roy, T.~Wald, G.~Koehler, M.~R. Rokuss, N.~Disch, J.~Holzschuh, D.~Zimmerer, and K.~H. Maier-Hein, ``Sam. md: Zero-shot medical image segmentation capabilities of the segment anything model,'' \emph{arXiv preprint arXiv:2304.05396}, 2023.

\bibitem{huang2024segment}
Y.~Huang, X.~Yang, L.~Liu, H.~Zhou, A.~Chang, X.~Zhou, R.~Chen, J.~Yu, J.~Chen, C.~Chen \emph{et~al.}, ``Segment anything model for medical images?'' \emph{Medical Image Analysis}, vol.~92, p. 103061, 2024.

\bibitem{ma2024segment}
J.~Ma, Y.~He, F.~Li, L.~Han, C.~You, and B.~Wang, ``Segment anything in medical images,'' \emph{Nature Communications}, vol.~15, no.~1, p. 654, 2024.

\bibitem{cen2023segment}
J.~Cen, Z.~Zhou, J.~Fang, W.~Shen, L.~Xie, D.~Jiang, X.~Zhang, Q.~Tian \emph{et~al.}, ``Segment anything in 3d with nerfs,'' \emph{Advances in Neural Information Processing Systems}, vol.~36, pp. 25\,971--25\,990, 2023.

\bibitem{yang2023sam3d}
Y.~Yang, X.~Wu, T.~He, H.~Zhao, and X.~Liu, ``Sam3d: Segment anything in 3d scenes,'' \emph{arXiv preprint arXiv:2306.03908}, 2023.

\bibitem{zhou2024point}
Y.~Zhou, J.~Gu, T.~Y. Chiang, F.~Xiang, and H.~Su, ``Point-sam: Promptable 3d segmentation model for point clouds,'' \emph{arXiv preprint arXiv:2406.17741}, 2024.

\bibitem{gong20233dsam}
S.~Gong, Y.~Zhong, W.~Ma, J.~Li, Z.~Wang, J.~Zhang, P.-A. Heng, and Q.~Dou, ``3dsam-adapter: Holistic adaptation of sam from 2d to 3d for promptable medical image segmentation,'' \emph{arXiv preprint arXiv:2306.13465}, 2023.

\bibitem{yu2023inpaint}
T.~Yu, R.~Feng, R.~Feng, J.~Liu, X.~Jin, W.~Zeng, and Z.~Chen, ``Inpaint anything: Segment anything meets image inpainting,'' \emph{arXiv preprint arXiv:2304.06790}, 2023.

\bibitem{zhang2023comprehensive}
C.~Zhang, L.~Liu, Y.~Cui, G.~Huang, W.~Lin, Y.~Yang, and Y.~Hu, ``A comprehensive survey on segment anything model for vision and beyond,'' 2023.

\bibitem{cheng2023segment}
Y.~Cheng, L.~Li, Y.~Xu, X.~Li, Z.~Yang, W.~Wang, and Y.~Yang, ``Segment and track anything,'' \emph{arXiv preprint arXiv:2305.06558}, 2023.

\bibitem{yang2023track}
J.~Yang, M.~Gao, Z.~Li, S.~Gao, F.~Wang, and F.~Zheng, ``Track anything: Segment anything meets videos,'' \emph{arXiv preprint arXiv:2304.11968}, 2023.

\bibitem{kweon2024sam}
H.~Kweon and K.-J. Yoon, ``From sam to cams: Exploring segment anything model for weakly supervised semantic segmentation,'' in \emph{Proceedings of the IEEE/CVF Conference on Computer Vision and Pattern Recognition}, 2024, pp. 19\,499--19\,509.

\bibitem{ruder2017knowledge}
S.~Ruder, P.~Ghaffari, and J.~G. Breslin, ``Knowledge adaptation: Teaching to adapt,'' \emph{arXiv preprint arXiv:1702.02052}, 2017.

\bibitem{he2019knowledge}
T.~He, C.~Shen, Z.~Tian, D.~Gong, C.~Sun, and Y.~Yan, ``Knowledge adaptation for efficient semantic segmentation,'' in \emph{Proceedings of the IEEE/CVF Conference on Computer Vision and Pattern Recognition}, 2019, pp. 578--587.

\bibitem{ma2014knowledge}
Z.~Ma, Y.~Yang, N.~Sebe, and A.~G. Hauptmann, ``Knowledge adaptation with partiallyshared features for event detectionusing few exemplars,'' \emph{IEEE transactions on pattern analysis and machine intelligence}, vol.~36, no.~9, pp. 1789--1802, 2014.

\bibitem{wang2021knowledge}
L.~Wang and K.-J. Yoon, ``Knowledge distillation and student-teacher learning for visual intelligence: A review and new outlooks,'' \emph{IEEE transactions on pattern analysis and machine intelligence}, vol.~44, no.~6, pp. 3048--3068, 2021.

\bibitem{rong2023boundary}
S.~Rong, B.~Tu, Z.~Wang, and J.~Li, ``Boundary-enhanced co-training for weakly supervised semantic segmentation,'' in \emph{Proceedings of the IEEE/CVF Conference on Computer Vision and Pattern Recognition}, 2023, pp. 19\,574--19\,584.

\bibitem{chen2023semantic}
J.~Chen, Z.~Yang, and L.~Zhang, ``Semantic segment anything,'' \url{https://github.com/fudan-zvg/Semantic-Segment-Anything}, 2023.

\bibitem{chen2023segment}
T.~Chen, Z.~Mai, R.~Li, and W.-l. Chao, ``Segment anything model (sam) enhanced pseudo labels for weakly supervised semantic segmentation,'' \emph{arXiv preprint arXiv:2305.05803}, 2023.

\bibitem{krishna1999genetic}
K.~Krishna and M.~N. Murty, ``Genetic k-means algorithm,'' \emph{IEEE Transactions on Systems, Man, and Cybernetics, Part B (Cybernetics)}, vol.~29, no.~3, pp. 433--439, 1999.

\bibitem{romera2017erfnet}
E.~Romera, J.~M. Alvarez, L.~M. Bergasa, and R.~Arroyo, ``Erfnet: Efficient residual factorized convnet for real-time semantic segmentation,'' \emph{IEEE Transactions on Intelligent Transportation Systems}, vol.~19, no.~1, pp. 263--272, 2017.

\bibitem{yue2021prototypical}
X.~Yue, Z.~Zheng, S.~Zhang, Y.~Gao, T.~Darrell, K.~Keutzer, and A.~S. Vincentelli, ``Prototypical cross-domain self-supervised learning for few-shot unsupervised domain adaptation,'' in \emph{Proceedings of the IEEE/CVF Conference on Computer Vision and Pattern Recognition}, 2021, pp. 13\,834--13\,844.

\bibitem{yang2021context}
K.~Yang, X.~Hu, and R.~Stiefelhagen, ``Is context-aware cnn ready for the surroundings? panoramic semantic segmentation in the wild,'' \emph{IEEE Transactions on Image Processing}, vol.~30, pp. 1866--1881, 2021.

\bibitem{armeni2017joint}
I.~Armeni, S.~Sax, A.~R. Zamir, and S.~Savarese, ``Joint 2d-3d-semantic data for indoor scene understanding,'' \emph{arXiv preprint arXiv:1702.01105}, 2017.

\bibitem{macpherson2022360}
I.~MacPherson, R.~F. Murray, and M.~S. Brown, ``A 360° omnidirectional photometer using a ricoh theta z1,'' in \emph{Color and Imaging Conference}, vol.~30.\hskip 1em plus 0.5em minus 0.4em\relax Society for Imaging Science and Technology, 2022, pp. 124--128.

\end{thebibliography}

\end{document}